\documentclass[lettersize,journal]{IEEEtran}
\usepackage{amsmath,amsfonts}
\usepackage{algorithmic}
\usepackage{algorithm}
\usepackage{enumitem}
\usepackage{array}
\usepackage{color}
\usepackage{textcomp}
\usepackage{stfloats}
\usepackage{url}
\usepackage{verbatim}
\usepackage{graphicx}
\usepackage{cite}
\usepackage{times}
\usepackage{soul}
\usepackage[table]{xcolor}
\usepackage{amsmath}
\usepackage{amssymb}
\usepackage{amsthm}
\usepackage{booktabs}
\usepackage{algorithm}
\usepackage{multirow}
\usepackage{subfigure}
\usepackage{tabularx}
\usepackage{subcaption}
\usepackage[switch]{lineno}

\usepackage{xcolor}

\begin{document}

\title{TERGAD: Structure-Aware Text-Enhanced Representations for Graph Anomaly Detection}

\author{Wen Shi,
Zhe Wang,
Huafei Huang,~\IEEEmembership{Graduate Student Member,~IEEE,}
Qing Qing,
Ziqi Xu,~\IEEEmembership{Member,~IEEE,}
Qixin Zhang,
Xikun Zhang,
Renqiang Luo,~\IEEEmembership{Member,~IEEE,}
Feng Xia,~\IEEEmembership{Fellow,~IEEE} 
\thanks{Shi Wen, Zhe Wang, Qing Qing, and Renqiang Luo are with College of Computer Science and Technology, Jilin University, Changchun 130012, China (\{shiwen24, qingqing25\}@mails.jlu.edu.cn, \{wz2000, lrenqiang\}@jlu.edu.cn).}
\thanks{Huafei Huang is with the School of Computer Science and Information Technology, Adelaide University, Adelaide SA5095, Australia (e-mail: hhuafei@outlook.com).}
\thanks{Ziqi Xu, Xikun Zhang, and Feng Xia are with the School of Computing Technologies, RMIT University, Melbourne, VIC 3000, Australia (e-mail: \{ziqi.xu, xikun.zhang\}@rmit.edu.au, f.xia@ieee.org).}
\thanks{Qixin Zhang is with College of Computing and Data Science, Nanyang Technological University, 639798, Singapore (e-mail: qixin.zhang2026@gmail.com).}
\thanks{Corresponding author: Xikun Zhang, Renqiang Luo.}}

\markboth{IEEE Transactions on Knowledge and Data Engineering,~Vol.~0, No.~0, May~2026}%
{Shell \MakeLowercase{\textit{et al.}}: A Sample Article Using IEEEtran.cls for IEEE Journals}


\maketitle

\begin{abstract}
Graph Anomaly Detection (GAD) aims to identify atypical graph entities, such as nodes, edges, or substructures, that deviate significantly from the majority. While existing text-rich approaches typically integrate structural context into the data representation pipeline using raw textual features, they often neglect the structural context of nodes. This limitation hinders their ability to detect sophisticated anomalies arising from inconsistencies between a node’s inherent content and its topological role. To bridge this gap, we propose TERGAD (Structure-aware Text-enhanced Representations for Graph Anomaly Detection), A novel data augmentation framework that enriches structural semantics for GAD via the semantic reasoning capabilities of Large Language Models.(LLMs). Specifically, TERGAD translates node-level topological properties into descriptive natural language narratives, which are subsequently processed by an LLM to derive high-level semantic embeddings. These embeddings are then adaptively fused with original node attributes through a gated dual-branch autoencoder to jointly reconstruct both graph structure and node features. The anomaly score is computed based on the integrated reconstruction error, effectively capturing deviations in both observable attributes and LLM-informed semantic expectations. Extensive experiments on six real-world datasets demonstrate that TERGAD consistently outperforms state-of-the-art baselines. Furthermore, our ablation studies validate the indispensable role of structural semantic guidance and the efficacy of the gated fusion mechanism. Code is available at~\url{https://github.com/Kantorakitty/TERGAD-main}.
\end{abstract}


\section{Introduction}
\IEEEPARstart{G}{raph} Anomaly Detection (GAD) aims to identify atypical graph objects, such as nodes, edges, or substructures, that deviate significantly from majority patterns in graph databases~\cite{zheng2024correlation}. 
This task has become indispensable in high-stakes domains, including financial fraud prevention~\cite{ma2023towards}, cyber intrusion detection~\cite{zhang2023federated}, and social network spam filtering~\cite{xiao2023counterfactual}. 
The growing interconnectedness of real-world data and rapid advancements in graph data mining have significantly boosted interest in GAD over the past decade~\cite{pazho2024a}. 
A key shift has been the transition from reliance on human expertise to traditional machine learning, and more recently, to sophisticated deep learning techniques~\cite{guo2023regraphgan}. 
These deep learning approaches enhance detection performance by learning complex, non-linear patterns directly from massive datasets through end-to-end data processing pipelines without manual feature engineering~\cite{qiao2025deep}.

\par In many real-world scenarios, however, graphs are ``text-poor" or purely structural, consisting only of nodes and edges without rich descriptive attributes. 
This lack of explicit semantic information poses a significant challenge, as anomalous nodes often hide within subtle structural patterns that are difficult to distinguish using numerical features alone~\cite{xia2026graph, luo2026utility}. 
Reconstruction-based analysis has become a common approach for detecting these anomalies, identifying nodes that the model fails to reconstruct accurately from learned representations. 
However, existing methods often struggle to interpret the functional meaning of a node's position within the data representation pipeline, as they rely on raw adjacency matrices or basic degree statistics~\cite{zhang2025survey}.
This limitation stems from an inability to translate abstract topological patterns into semantically rich representations that provide a higher-level understanding of node behavior. 
For example, a node with a high clustering coefficient might be a normal hub in one community but a malicious bridge in another, a nuance that numerical feature engineering pipelines often miss.

\par
\begin{figure}[t]
    \centering
    \includegraphics[width=0.95\columnwidth]{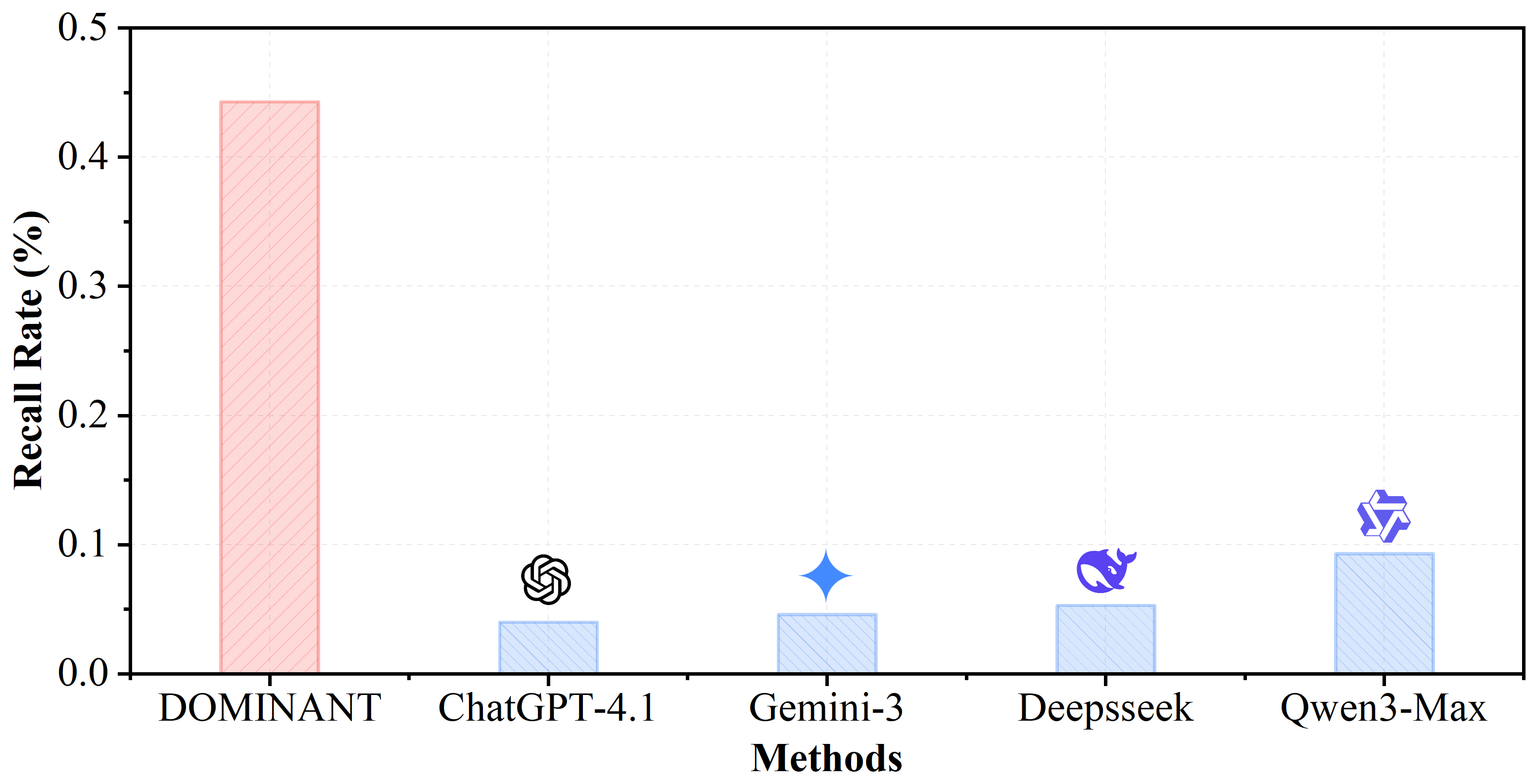}
    \caption{Performance comparison between DOMINANT, a traditional GCN-based method, and four representative LLMs. The significant performance gap motivates the need for the integrated semantic-structural approach in LLM-based GAD.}
    \label{fig:bac}
\end{figure}

 To overcome the sparsity of node attributes, we explore the potential of transforming structural patterns into text-rich representations via Large Language Models (LLMs) as a data augmentation strategy~\cite{luo2026fairge}. 
A growing paradigm involves converting graph topology into natural language descriptions, allowing the model to leverage the vast reasoning capabilities of LLMs for structural understanding. 
For instance, recent works have attempted to prompt LLMs to produce explanations or labels based on graph connectivity~\cite{he2024harnessing, guo2023gpt4graph}. 
These approaches demonstrate promising results in tasks that benefit from natural language reasoning and external knowledge integration~\cite{luo2026bridging}. 
However, standard Graph2Text methods suffer from critical limitations, such as requiring LLMs to infer implicit structures from long, sequential text descriptions. 
Furthermore, these methods often produce excessively long sequences that exceed the practical context windows of most LLMs, making them unscalable for large-scale graph data. 
Moreover, most current methods lack an explicit mechanism to balance these newly generated semantic priors with the original structural properties of the graph within a unified feature integration framework.
To further investigate the direct applicability of LLMs in this domain, we conducted a preliminary study evaluating vanilla LLMs on the GAD task. 
The results (in Fig.~\ref{fig:bac}) reveal their zero-shot performance significantly lags behind GCN-based GAD models, suggesting that LLMs struggle to inherently grasp the underlying structural anomalies without specialized adaptation. 
These findings underscore the non-trivial nature of effectively leveraging LLMs for graph anomaly detection and motivate the need for a more integrated semantic-structural approach.

\par Addressing these gaps leads to a fundamental question: \textbf{Can we effectively empower structural graph learning by synthesizing text-rich representations through LLMs?} 
Answering this question requires overcoming several technical hurdles regarding efficiency and integration. 
First, we must determine how to effectively translate abstract graph topology into semantically rich yet computationally efficient textual descriptions within a structured data transformation pipeline. 
Second, it is crucial to integrate these LLM-derived semantic insights with specialized graph learning models in a complementary manner. 
Third, we need to design a framework that jointly optimizes both the synthesized semantic consistency and the native structural information for the anomaly detection task. 
Effectively balancing these multi-modal signals is essential for capturing the subtle deviations that characterize sophisticated anomalies in otherwise attribute-poor networks.

\par Our work addresses these challenges by establishing a vital connection between raw structural information and LLM-based semantic understandings within a unified data processing framework. 
We propose \textbf{TERGAD}, a novel framework for structure-aware \textbf{T}ext-\textbf{E}nhanced \textbf{R}epresentations for \textbf{G}raph \textbf{A}nomaly \textbf{D}etection, which establishes a vital connection between raw structural information and LLM-based semantic reasoning. 
Specifically, our framework enriches the representation of each node by generating natural language descriptions that articulate its unique structural role within the network. 
This is achieved through carefully designed templates that translate complex topological metrics, such as centrality and community membership, into concise and interpretable textual roles. 
A gated dual-branch autoencoder then adaptively fuses these synthesized semantic embeddings with the original structural features through a learnable gating mechanism. 
This architecture jointly reconstructs both the synthesized semantic descriptions and the underlying graph structure, with anomaly scores reflecting deviations in both domains.

Our main contributions are summarized as follows:
\begin{itemize} [leftmargin=0.5cm]
    \item \textbf{Constructing Text-Rich Graphs for Anomaly Detection.} 
    We propose TERGAD, a framework that addresses attribute sparsity by integrating explicit structural semantics derived from LLMs through a systematic semantic enrichment pipeline. 
    By translating topology into natural language, we transform purely structural graphs into text-rich representations for more robust detection.
    
    \item \textbf{Semantic-Guided Graph Encoding and Adaptive Fusion.} 
    We introduce a structured template-based approach to encode topological roles via LLMs, followed by an adaptive fusion mechanism. 
    This allows the model to selectively integrate high-level semantic priors with raw structural data based on the specific context of each node.
    
    \item \textbf{Extensive Empirical Validation.} 
    We conduct comprehensive experiments on six real-world datasets to evaluate the effectiveness of our proposed framework within practical data engineering. 
    The results demonstrate that TERGAD consistently outperforms state-of-the-art GAD methods.
\end{itemize}

\section{Related Work}
\subsection{LLMs for Graph Learning}
\par Recent advances in LLMs have spurred growing interest in applying their reasoning capabilities to graph~\cite{zhang2025can}. 
A dominant paradigm involves converting graph topology into natural language descriptions through structured data transformation pipelines (often termed Graph2Text), which are then processed by frozen LLMs to generate semantic node features~\cite{zhang2025taga}. 
For instance, a unified vandalism detection system for Wikidata employs a Graph2Text approach to convert complex factual triples and multilingual edits into a single textual space, allowing a language model to evaluate both structural and content changes for potential knowledge alterations~\cite{trokhymovych2025graph}. 
Similarly, GPT4Graph~\cite{guo2023gpt4graph} systematically evaluates the structural understanding of LLMs by translating graphs into various textual formats like adjacency lists and GraphML.

\par However, the Graph2Text approach suffers from two critical limitations in data processing efficiency: it forces LLMs to infer implicit graph structures from sequential text, which is inefficient compared to native graph learners, and often produces excessively long sequences that exceed LLM context windows. 
To address these issues, InstructGLM~\cite{ye2024language} proposes instruction-finetuning LLMs directly on graph tasks using scalable natural language prompts that explicitly describe multi-hop neighborhoods. 
In a different vein, GraphLLM~\cite{chai2025graphllm} integrates a graph transformer with an LLM via graph-enhanced prefix tuning, condensing graph information into a compact prefix and bypassing verbose conversion.

\par Despite these innovations, most existing methods either treat LLMs as static feature extractors or require complex multi-stage training pipelines. 
Moreover, the generated representations often lack explicit mechanisms to balance original attribute fidelity and LLM-derived semantic priors. 
Our work, TERGAD, directly addresses this gap through a gated dual-branch architecture that adaptively fuses both modalities for robust anomaly detection within real-world data quality assessment workflows.

\subsection{Graph Anomaly Detection}
\par Various deep learning-based approaches have been developed to tackle the challenges of GAD. 
Graph Auto-Encoders (GAEs) serve as a foundational paradigm, identifying anomalies by measuring the reconstruction quality of node embeddings. 
For instance, AnomalyDAE~\cite{fan2020anomalydae} employs a dual auto-encoder architecture with attention mechanisms to capture the complex interplay between network topology and node attributes.
GAAN~\cite{chen2020generative} introduces a generative adversarial framework, integrating a generator and discriminator to detect anomalies through both reconstruction errors and discrimination confidence.

\par Beyond reconstruction, self-supervised learning has gained significant traction. 
CoLA~\cite{liu2021anomaly} utilizes contrastive learning on instance pairs to enhance scalability, while CONAD~\cite{xu2022contrastive} incorporates anomaly-specific prior knowledge via data augmentation and a Siamese GNN encoder. 
More recent works focus on local contexts and data scarcity; GAD-NR~\cite{roy2024gad} models local structures through neighborhood reconstruction, and FIAD~\cite{chen2025fiad} alleviates label scarcity by injecting synthetic anomaly signals directly into the feature matrix. 
To address multi-modal inconsistencies, AHFAN~\cite{wang2025graph} employs an attention-based module to fuse topology-driven and semantics-driven representations.

\par Despite these advancements, most existing GAD methods operate strictly within numerical topological and attribute spaces. 
They lack the capacity to incorporate high-level semantic knowledge regarding graph structural data, such as the functional roles nodes play, which limits their detection performance and interpretability on complex, real-world datasets. 
This semantic void  in data representation motivates our exploration of LLMs to bridge the gap between raw structural data and high-level conceptual understanding.

\section{Preliminaries}

\subsection{Notations}
\par Unless otherwise specified, we adopt the following mathematical notations throughout this paper: 
sets are denoted by calligraphic uppercase letters (e.g., $\mathcal{A}$), matrices by bold uppercase letters (e.g., $\mathbf{A}$), and vectors by bold lowercase letters (e.g., $\mathbf{x}$).

\par We define a graph as $\mathcal{G} = (\mathcal{V}, \mathcal{E}, \mathbf{X})$, where $\mathcal{V}$ represents the set of $n$ nodes ($ |\mathcal{V}|= n$), $\mathcal{E} \subseteq \mathcal{V} \times \mathcal{V}$ is the set of edges, and $\mathbf{X} \in \mathbb{R}^{n \times d_x}$ denotes the node attribute matrix. 
The topological structure is represented by an adjacency matrix $\mathbf{A} \in \{0,1\}^{n \times n}$, where $\mathbf{A}[i,j] = 1$ if an edge exists between node $i$ to node $j$, and 0 otherwise.
To facilitate graph learning, we compute the normalized adjacency matrix $\tilde{\mathbf{A}} = \mathbf{D}^{-1/2}(\mathbf{A} + \mathbf{I})\mathbf{D}^{-1/2}$, where $\mathbf{I}$ is the identity matrix and $\mathbf{D}$ is the degree matrix of $\mathbf{A} + \mathbf{I}$.

\par Furthermore, we let $\mathbf{Z}_{\text{LLM}} \in \mathbb{R}^{n \times d_z}$ denote the semantic embeddings generated by LLMs from natural language descriptions of the graph. 
Our dual-branch encoder maps input features into two latent representations: 
$\mathbf{H}^{(x)} \in \mathbb{R}^{n \times h}$ from the attribute branch and $\mathbf{H}^{(z)} \in \mathbb{R}^{n \times h}$ from the LLM branch, with $h$ being the hidden dimension. 
These are integrated via an adaptive gate matrix $\mathbf{G} \in [0,1]^{n \times h}$ to obtain the final fused representation $\mathbf{Z} \in \mathbb{R}^{n \times h}$. 
The decoder then reconstructs the node attributes and adjacency matrix, denoted as $\hat{\mathbf{X}}$ and $\hat{\mathbf{A}}$ respectively. 
For each node $i$, the anomaly score is computed as $s_i$, where $\alpha \in [0,1]$ is a hyperparameter balancing the respective reconstruction losses.
A complete summary of notations is provided in Table~\ref{tab:notations}.

\begin{table}[!t]
\centering
\footnotesize
    \renewcommand{\arraystretch}{1.2}
\caption{Notations and their definitions.}
\label{tab:notations}
\begin{tabularx}{\columnwidth}{>{\raggedright\arraybackslash}l >{\raggedright\arraybackslash}X}
\toprule
\textbf{Notation} & \textbf{Definition} \\
\midrule
$\mathcal{G}$ & Graph \\
$\mathcal{V}$ & The set of nodes \\
$\mathcal{E}$ & The set of edges \\
$\mathbf{A}$ & Adjacency matrix \\
$\tilde{\mathbf{A}}$ & Normalized adjacency matrix \\
$\mathbf{X}$ & Node attribute matrix \\
$\mathbf{Z}_{\text{LLM}}$ & LLM-derived semantic embedding matrix \\
$\mathbf{G}$ & Gate matrix for adaptive fusion \\
$\mathbf{Z}$ & Final fused latent representation \\
$\hat{\mathbf{X}}$ & Reconstructed node attributes \\
$\hat{\mathbf{A}}$ & Reconstructed adjacency matrix \\
$d_x$ & The number of node attributes \\
$d_z$ & Dimension of LLM embeddings \\
$s_i$ & Anomaly score for node $V_i$ \\
\bottomrule
\end{tabularx}
\end{table}

\subsection{Graph Anomaly Detection}
\par Given an attributed graph $\mathcal{G} = (\mathcal{V}, \mathcal{E}, \mathbf{X})$,
The goal of GAD is to identify a subset of nodes \( \mathcal{V}_{\text{anom}} \subset \mathcal{V} \) that exhibit significant deviations from the majority in terms of structural connectivity or attribute patterns.
A prevalent approach in this domain leverages Graph Autoencoders based on Graph Convolutional Networks (GCNs). 
In this framework, the GCN encoder utilizes graph convolution layers to learn node representations by aggregating information from local neighborhoods.
A GCN layer is defined as:

\begin{equation}
    \mathbf{H}^{(l+1)} = \sigma\left(\tilde{\mathbf{A}} \mathbf{H}^{(l)} W^{(l)}\right),
\end{equation}
where $\tilde{\mathbf{A}} = \mathbf{D}^{-\frac{1}{2}}(\mathbf{A} + \mathbf{I})\mathbf{D}^{-\frac{1}{2}}$ is the normalized adjacency matrix with self-loops, $\mathbf{H}^{(l)}$ is the node representations at layer $l$, $\mathbf{W}^{(l)}$ is the trainable weight matrix, and $\sigma$ denotes an activation function.

\par The encoder maps nodes to their latent representations, \textit{i.e.} $\mathbf{Z} = \text{Encoder}(\mathbf{X}, \mathbf{A})$, and the decoder is adopted to reconstruct the original graph information, typically the adjacency matrix: 
$\hat{\mathbf{A}} = \text{Decoder}(\mathbf{Z})$. 
By minimizing the reconstruction error, the structural patterns of the graph are encoded in the GCN parameters. 
A common reconstruction error formulation is:

\begin{equation}
\mathcal{L}_{\text{struct}} = \|\mathbf{A} - \hat{\mathbf{A}}\|_\text{F}^2.
\end{equation}

\par For attributed graphs, the attribute reconstruction loss may also be included:

\begin{equation}
\mathcal{L}_{\text{attr}} = \|\mathbf{X} - \hat{\mathbf{X}}\|_F^2.
\end{equation}

\par The final anomaly score $s_i$ for a node $v_i$ is derived from its combined reconstruction errors, reflecting its deviation from the learned normal pattern. A typical scoring function is:

\begin{equation}
    s_i = (1 - \alpha) \, \|\mathbf{a}_i - \hat{\mathbf{a}}_i\|_2^2 + \alpha \, \|\mathbf{x}_i - \hat{\mathbf{x}}_i\|_2^2,
\end{equation}
where $\mathbf{a}_i$ and $\hat{\mathbf{a}}_i$ are the original and reconstructed adjacency vectors for node $i$, $\mathbf{x}_i$ and $\hat{\mathbf{x}}_i$ are its original and reconstructed attribute vectors, and $\alpha \in [0,1]$ is a hyperparameter balancing the two terms. 
Nodes with higher scores $s_i$ are considered more anomalous.

\begin{figure*}[t]
    \centering
    \includegraphics[width=0.9\textwidth]{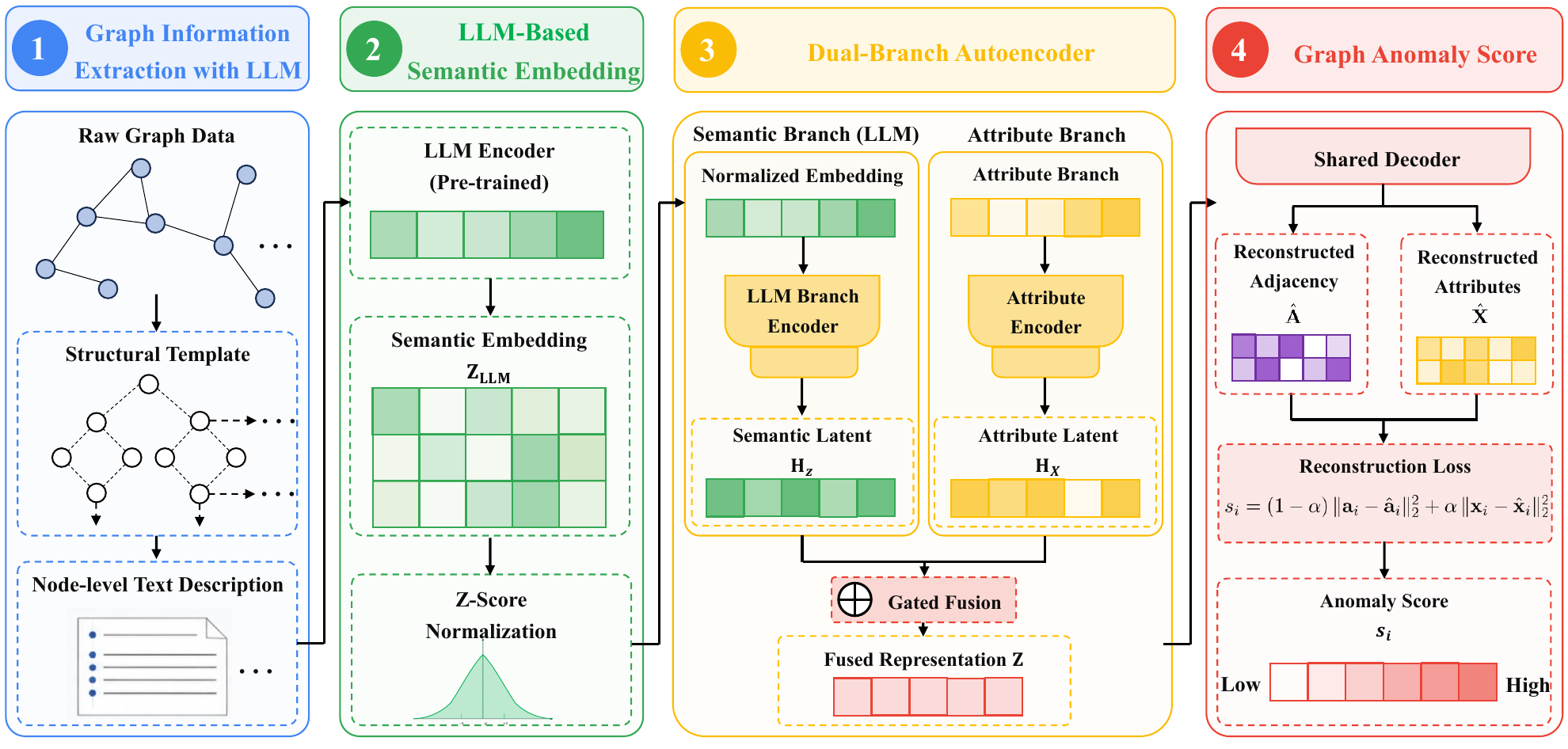}
    \caption{The architecture of TERGAD.}
    \label{fig:llmforgad-architecture}
\end{figure*}

\section{The Design of TERGAD}
\par In this section, we introduce the architecture of TERGAD (Fig. \ref{fig:llmforgad-architecture}). 
We first present the graph information extraction with LLMs and the LLM-based semantic embedding, followed by the design of the dual-branch autoencoder. 
Finally, we define the anomaly score based on TERGAD.
Our algorithm is summarized in Algorithm~\ref{alg:TERGAD}.

\subsection{Graph Structure Data Extraction with LLMs}
\label{sec:prompt}
\par To bridge the modality gap between graph-structured data and the textual inputs required by LLMs, we transform raw graph inputs into node-level natural language descriptions via a structured prompting template. 
This process begins with a comprehensive JSON-formatted data intermediate representation that encodes global statistics (e.g., node/edge counts, directedness), degree distributions, high-order topological features (e.g., clustering coefficients, $k$-core numbers, centrality measures), and community structures.

\par Our template dynamically constructs descriptive sentences by contextualizing raw values into human-interpretable semantics:
\begin{itemize} [leftmargin=0.5cm]
    \item \textbf{Connectivity}: Degree is described relative to global percentiles (e.g., ``top $1$\% of all nodes by degree") or absolute ranks to signify its relative importance.
    \item \textbf{Local Cohesion}: The clustering coefficient is mapped to qualitative labels (e.g., ``high", ``moderate") while retaining precise numerical values for granularity.
    \item \textbf{Global Significance}: Centrality metrics are annotated with percentile-based significance (e.g., ``critical bridge node within the top $10$\%").
    \item \textbf{Topological Roles}: Structural roles (e.g., hub, peripheral node, core member) are explicitly assigned based on multi-criteria heuristic thresholds.
\end{itemize}
    
\par
\begin{figure}[t]
    \centering
    \includegraphics[width=\linewidth]{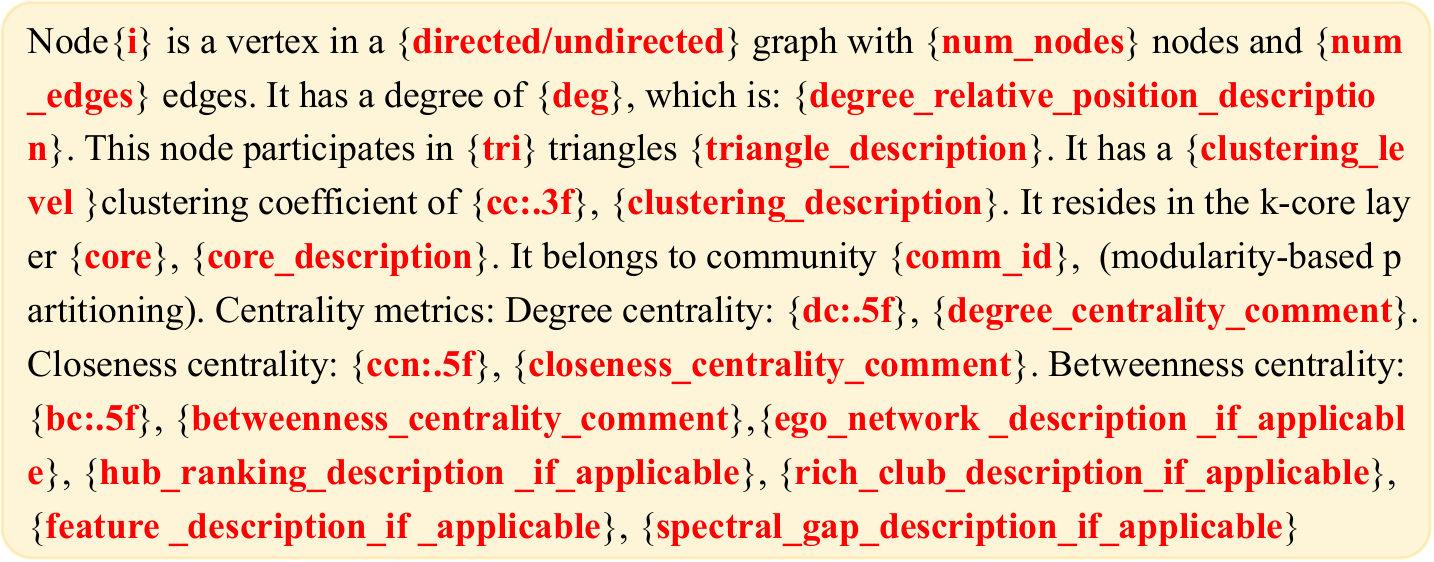}
    \caption{Structured text description template for each node.}
    \label{fig:prompt_figure}
\end{figure}

 Each description is initialized with a base identity sentence: ``Node $i$ is a vertex in an undirected graph...". 
It is crucial to note that the node index ($i$) serves exclusively as a persistent identifier within the current graph instance to maintain cross-modal alignment between the structural branch and the LLM branch. 
These IDs carry no global semantic meaning and are not utilized as features for prediction, thereby ensuring that no test-set information is leaked and the model's generalizability is preserved. 
This design ensures that the resulting text is fluent, self-contained, and semantically aligned with the reasoning patterns of off-the-shelf LLMs without requiring task-specific fine-tuning. 
A detailed illustration of the template is shown in Fig.~\ref{fig:prompt_figure}.

\par \textbf{Illustrative Example}: Node $42$ is a vertex in an undirected graph with $6,000$ nodes and $8,000$ edges. 
It has a degree of $1,248$, placing it in the top $1$\% of all nodes by connectivity. 
This node participates in $3,892$ triangles, indicating its immersion in tightly-knit local clusters. 
With a local clustering coefficient of $0.756$, nearly all its neighbors are interconnected. 
It resides in the $k$-core layer $12$, remaining connected even after the iterative removal of lower-degree nodes, which identifies it as a structurally robust core member. 
Furthermore, it belongs to community $3$. 
Centrality Analysis: Degree centrality ($0.852$, top $10$\%) indicates high direct influence; 
Closeness centrality ($0.723$, top $10$\%) reflects its proximity to the network center; 
Betweenness centrality ($0.612$, top $10$\%) marks it as a critical bridge. 
In its $1$-hop ego network, it maintains $1248$ neighbors with an average degree of $45.32 \pm 12.45$. 
This node is among the top $3$ primary hubs. 
Global spectral analysis shows a Fiedler value of $0.154$, indicating strong algebraic connectivity across the graph.

\subsection{LLM-Based Semantic Embedding}
\label{subsec:LLM-Based Semantic Embedding}
\par We encode the generated natural language descriptions into semantic embeddings using the \textbf{BGE-large-en-v$1.5$}. 
Following the architecture, we apply \textit{last-token pooling} to extract the hidden state of the last non-padding token as the sentence representation. 
To avoid interference with subsequent standardization and ensure effective statistical alignment, we disable the default L$2$ normalization.

\par To ensure compatibility with the original node attributes and stabilize downstream autoencoder training, we then apply \textbf{Z-score standardization} across all nodes for each embedding dimension:
\begin{equation}
    \mathbf{z}_i^{\text{norm}} = \frac{\mathbf{z}_i - \mu}{\sigma},
    \label{eq:zscore}
\end{equation}
where $\mu = \frac{1}{n}\sum_{j=1}^n \mathbf{z}_j$ and $\sigma = \sqrt{\frac{1}{n}\sum_{j=1}^n (\mathbf{z}_j - \mu)^2}$ denote the mean and standard deviation of the $i$-th embedding dimension over all $n$ nodes. 
This transformation aligns the statistical distribution of LLM-derived embeddings with typical preprocessed graph features, enabling balanced reconstruction in the dual-branch architecture. 
The final embeddings $\mathbf{Z}_{\text{LLM}} \in \mathbb{R}^{n \times 1024}$ serve as enhanced attribute inputs to the dual-branch autoencoder.

\subsection{Dual-Branch Autoencoder}
\par Our framework begins with two complementary inputs: the original node attributes $\mathbf{X} \in \mathbb{R}^{n \times d_x}$ and the LLM-derived semantic embeddings $\mathbf{Z}_{\text{LLM}} \in \mathbb{R}^{n \times d_z}$, where the latter is standardized via Z-score normalization as described in Section~\ref{subsec:LLM-Based Semantic Embedding}. 
While $\mathbf{X}$ captures task-specific features (e.g., textual content or metadata), $\mathbf{Z}_{\text{LLM}}$ encodes high-level structural semantics inferred from natural language descriptions of the graph. 
By preserving both modalities as separate inputs, we avoid premature fusion and allow the model to learn modality-specific graph representations before integration.

\par The graph structure is encoded using a normalized adjacency matrix $\tilde{\mathbf{A}} = \mathbf{D}^{-1/2} (\mathbf{A} + \mathbf{I}) \mathbf{D}^{-1/2}$, following the standard GCN framework~\cite{kipf2017semi}. 
This preprocessing ensures stable message propagation and incorporates self-loops to retain node-specific information.

\par The encoder consists of two parallel graph convolutional branches:
\begin{align}
    \mathbf{H}^{(x)} &= \sigma\left( \tilde{\mathbf{A}} \mathbf{X} \mathbf{W}^{(x)} \right), \\
    \mathbf{H}^{(z)} &= \sigma\left( \tilde{\mathbf{A}} \mathbf{Z}_{\text{LLM}} \mathbf{W}^{(z)} \right),
\end{align}
where $\mathbf{W}^{(x)} \in \mathbb{R}^{d_x \times h}$ and $\mathbf{W}^{(z)} \in \mathbb{R}^{d_z \times h}$ are learnable projection matrices, and $\sigma(\cdot)$ denotes the ReLU activation function. 
Each branch independently aggregates neighborhood information within its own semantic space, thereby preserving the distinct characteristics of raw attributes and LLM-enhanced embeddings.
\begin{algorithm}[t]
\footnotesize
\caption{TERGAD: Graph Anomaly Detection via LLM-Augmented Semantic Fusion}
\label{alg:TERGAD}
\begin{algorithmic}[1]
\STATE \textbf{Input:} Graph $\mathcal{G} = (\mathcal{V}, \mathbf{A}, \mathbf{X})$, natural language template $\mathcal{T}$, LLM $\mathcal{M}_{\text{LLM}}$, hidden dimension $h$, loss weight $\alpha$
\STATE \textbf{Output:} Anomaly scores $\{s_i\}_{i=1}^n$ for all nodes
\STATE
\STATE \textbf{Phase 1: Graph-to-Text Conversion}
\FOR{each node $v_i \in \mathcal{V}$}
    \STATE $t_i \gets \mathcal{T}(v_i, \mathcal{N}(v_i), \mathbf{X})$
\ENDFOR
\STATE $\mathbf{T} \gets [t_1, t_2, \dots, t_n]$
\STATE
\STATE \textbf{Phase 2: LLM Semantic Embedding}
\STATE $\mathbf{Z}_{\text{LLM}} \gets \mathcal{M}_{\text{LLM}}(\mathbf{T})$
\STATE $\mathbf{Z}_{\text{LLM}} \gets \operatorname{StandardScaler}(\mathbf{Z}_{\text{LLM}})$
\STATE
\STATE \textbf{Phase 3: Dual-Branch Gated Autoencoder}
\STATE $\tilde{\mathbf{A}} \gets \mathbf{D}^{-1/2}(\mathbf{A} + \mathbf{I})\mathbf{D}^{-1/2}$
\STATE $\mathbf{H}^{(x)} \gets \operatorname{ReLU}(\tilde{\mathbf{A}} \mathbf{X} \mathbf{W}^{(x)})$
\STATE $\mathbf{H}^{(z)} \gets \operatorname{ReLU}(\tilde{\mathbf{A}} \mathbf{Z}_{\text{LLM}} \mathbf{W}^{(z)})$
\STATE $\mathbf{G} \gets \sigma\left( \mathbf{W}_g [\mathbf{H}^{(x)} \parallel \mathbf{H}^{(z)}] + \mathbf{b}_g \right)$
\STATE $\mathbf{H}^{\text{fused}} \gets \mathbf{G} \odot \mathbf{H}^{(x)} + (1 - \mathbf{G}) \odot \mathbf{H}^{(z)}$
\STATE $\mathbf{Z} \gets \operatorname{ReLU}(\tilde{\mathbf{A}} \mathbf{H}^{\text{fused}} \mathbf{W}^{\text{fuse}})$
\STATE
\STATE \textbf{Phase 4: Reconstruction and Scoring}
\STATE $\hat{\mathbf{X}} \gets \tilde{\mathbf{A}} \mathbf{Z} \mathbf{W}^{\text{dec}}$
\STATE $\hat{\mathbf{A}} \gets \sigma(\mathbf{Z} \mathbf{Z}^\top)$
\FOR{each node $i = 1$ \textbf{to} $n$}
    \STATE $s_i \gets (1 - \alpha) \cdot \|\mathbf{a}_i - \hat{\mathbf{a}}_i\|_2^2 + \alpha \cdot \|\mathbf{x}_i - \hat{\mathbf{x}}_i\|_2^2$
\ENDFOR
\STATE return $\{s_i\}_{i=1}^n$
\end{algorithmic}
\end{algorithm}

\par To combine these representations in a context-aware manner, we introduce a gated fusion mechanism:
\begin{align}
    \mathbf{G} &= \sigma_g\left( \mathbf{W}_g [\mathbf{H}^{(x)} \parallel \mathbf{H}^{(z)}] + \mathbf{b}_g \right), \\
    \mathbf{H}^{\text{fused}} &= \mathbf{G} \odot \mathbf{H}^{(x)} + (1 - \mathbf{G}) \odot \mathbf{H}^{(z)},
\end{align}
where $\sigma_g$ is the sigmoid function, $\mathbf{W}_g \in \mathbb{R}^{h \times 2h}$ and $\mathbf{b}_g \in \mathbb{R}^{h}$ are trainable parameters, and $\odot$ denotes element-wise multiplication. 
The gate $\mathbf{G} \in [0,1]^{n \times h}$ dynamically determines the contribution of each branch for every node and feature dimension. 
This adaptability is essential: for nodes with reliable structural roles, LLM semantics may dominate; for ambiguous or sparse cases, raw attributes may carry more weight.

\par The fused representation is further refined through an additional graph convolutional layer:
\[
\mathbf{Z} = \sigma\left( \tilde{\mathbf{A}} \mathbf{H}^{\text{fused}} \mathbf{W}^{\text{fuse}} \right),
\]
with $\mathbf{W}^{\text{fuse}} \in \mathbb{R}^{h \times h}$. 
This step enables higher-order interactions among neighbors on the integrated signal, enhancing the expressiveness of the final latent representation.

\par Finally, a shared decoder reconstructs both the original node attributes and the graph structure from $\mathbf{Z}$:
\begin{align}
    \hat{\mathbf{X}} &= \tilde{\mathbf{A}} \mathbf{Z} \mathbf{W}^{\text{dec}}, \\
    \hat{\mathbf{A}} &= \sigma_s\left( \mathbf{Z} \mathbf{Z}^\top \right),
\end{align}
where $\mathbf{W}^{\text{dec}} \in \mathbb{R}^{h \times d_x}$ is a learnable weight matrix, and $\sigma_s(\cdot)$ is the sigmoid function that maps the reconstructed edge values into the $[0,1]$ range. 
Crucially, the attribute decoder reproduces the original attributes $\mathbf{X}$ rather than LLM embeddings, ensuring reconstructed features remain grounded in the observable data space.

\subsection{Graph Anomaly Score}
\par Within the TERGAD framework, the anomaly detection is performed via the reconstruction errors of the dual-branch autoencoder. 
For a given node $v_i$, the anomaly score is computed as:
\begin{equation}
    s_i = (1 - \alpha) \, \|\mathbf{a}_i - \hat{\mathbf{a}}_i\|_2^2 + \alpha \, \|\mathbf{x}_i - \hat{\mathbf{x}}_i\|_2^2,
    \label{eq:anomaly_score}
\end{equation}
where $\alpha \in [0, 1]$ balances the contributions of structural and attribute reconstruction errors. 
Here, $\mathbf{a}_i$ and $\hat{\mathbf{a}}_i$ denote the $i$-th rows of the ground-truth adjacency matrix $\mathbf{A}$ and its reconstructed version $\hat{\mathbf{A}}$, respectively, while $\mathbf{x}_i$ and $\hat{\mathbf{x}}_i$ are the $i$-th rows of the original node attributes $\mathbf{X}$ and the reconstructed $\hat{\mathbf{X}}$. 
Notably, $\hat{\mathbf{X}}$ is derived from a fused latent representation that integrates both the original attributes and LLM-enhanced semantic embeddings, allowing the model to leverage external semantic knowledge for more robust attribute reconstruction. 
Higher scores indicate greater deviation from normal patterns and thus higher anomaly likelihood.

\section{Experiments}
\subsection{Datasets}
\par In our experiments, we employ six widely adopted graph datasets.
Detailed statistics of these datasets are summarized in Table~\ref{tab:datastatistics}.
\begin{itemize} [leftmargin=0.5cm]
    \item \textbf{Cora}~\cite{sen2008collective} is a citation network consisting of $2,708$ machine learning papers grouped into $7$ categories. 
    The graph has $5,429$ citation edges, and node attributes are bag-of-words features derived from the paper abstracts.
    \item \textbf{Citeseer}~\cite{sen2008collective} is a citation network comprising $3,327$ scientific publications from the Citeseer digital library. 
    The graph contains $4,732$ citation edges, and node attributes are bag-of-words features based on the paper content.
    \item \textbf{DBLP}~\cite{yuan2021higher} is a citation network comprising $5,484$ publications from the DBLP Computer Science Bibliography, with $8,117$ citation edges. Node attributes are extracted from article titles.
    \item \textbf{ACM}~\cite{tang2008arnetminer} is a citation network with $16,484$ papers from the ACM digital library. 
    The graph has $71,980$ citation edges, and node attributes are bag-of-words features extracted from the paper titles and abstracts.
    \item \textbf{Pubmed}~\cite{sen2008collective} is a citation network of $19,717$ scientific publications from the PubMed database. 
    The graph includes $88,648$ citation links, and node attributes are TF-IDF weighted word frequencies from the paper abstracts.
    \item \textbf{BlogCatalog}~\cite{tang2009relational} is a social network dataset sourced from the BlogCatalog blogging platform (shortened to \textbf{Blog} in this paper).
    Nodes correspond to users, while edges capture the follower-followee relationships among them.
    Node features are derived from user-generated content, including blog posts and photo tags.

\end{itemize}

\begin{table} [t]
    \centering
    \footnotesize
    \renewcommand{\arraystretch}{1.2}
    \caption{The statistics of six real-world datasets.}
    \begin{tabular}{lrrrrl}
    \toprule
    \textbf{Dataset} & \textbf{\# Nodes} & \textbf{\# Edges} & \textbf{\# Feature} & \textbf{\# Anomaly} & \textbf{Ratio} \\ \midrule
    \textbf{Cora} & $2,708$ & $5,429$ & $1,433$ & $150$ & $5.54$\% \\
    \textbf{Citeseer} & $3,327$ & $4,732$ & $3,703$ & $150$ & $4.51$\% \\
    \textbf{DBLP} & $5,484$ & $8,117$ & $6,775$ & $273$ & $4.98$\% \\
    \textbf{ACM} & $16,484$ & $71,980$ & $8,337$ & $597$ & $3.62$\% \\
    \textbf{Pubmed} & $19,717$ & $88,648$ & $500$ & $600$ & $3.04$\% \\ 
    \textbf{Blog} & $5,196$ & $171,743$ & $8,189$ & $300$ & $5.77$\% \\
    \bottomrule
    \end{tabular}
    \label{tab:datastatistics}
\end{table}

\begin{table*}
    \centering
    \caption{Performance comparison of TERGAD and baselines across six datasets. Results are reported in using ROC-AUC ($\%$) and PR-AUC (\%), and the best performance in each column is highlighted in \textcolor{red}{\textbf{red and bolded}}.}
    \renewcommand{\arraystretch}{1.2}
    \tabcolsep=0.25cm
    \begin{tabular}{llcccccccc}
    \toprule
    Method & & Cora & Citeseer & DBLP & ACM & Pubmed  & BlogCatalog\\
    \midrule
    \rowcolor{green!20} \multicolumn{8}{c}{\textbf{ROC-AUC}} \\
    MLPAE &(MLSDA $'12$) & $76.06_{\pm 0.08}$ & $72.10_{\pm 0.04}$ & $74.32_{\pm 0.11}$ & $74.89_{\pm 0.04}$ & $74.56_{\pm 0.12}$  & $77.49_{\pm 0.11}$ \\
    GCNAE &(ICLR $'17$) & $76.05_{\pm 0.03}$ & $71.95_{\pm 0.14}$ & $74.40_{\pm 0.05}$ & $74.84_{\pm 0.09}$ & $74.41_{\pm 0.07}$  & $74.66_{\pm 0.10}$ \\
    DOMINANT &(SDM $'19$) & $88.52_{\pm 0.13}$ & $92.85_{\pm 0.10}$ & $95.41_{\pm 0.08}$ & $80.58_{\pm 0.04}$ & $81.40_{\pm 0.07}$ & $69.92_{\pm 0.06}$ \\
    AnomalyDAE &(ICASSP $'20$) & $76.27_{\pm 0.11}$ & $72.08_{\pm 0.05}$ & $74.17_{\pm 0.14}$ & $74.86_{\pm 0.06}$ & $74.79_{\pm 0.13}$ & $74.80_{\pm 0.14}$ \\
    CoLA &(TNNLS $'22$) & $67.62_{\pm 0.06}$ & $71.01_{\pm 0.10}$ & $55.67_{\pm 0.12}$ & $53.95_{\pm 0.13}$ & $56.29_{\pm 0.08}$ & $63.69_{\pm 0.09}$\\
    GAAN &(CIKM $'20$) & $76.79_{\pm 0.14}$ & $70.76_{\pm 0.08}$ & $73.93_{\pm 0.11}$ & $88.25_{\pm 0.05}$ & $90.35_{\pm 0.07}$ & $80.29_{\pm 0.03}$ \\
    GADNR &(WSDM $'24$) & $75.36_{\pm 0.13}$ & $75.04_{\pm 0.03}$ & $73.93_{\pm 0.10}$ & $71.72_{\pm 0.06}$ & $78.14_{\pm 0.04}$  & $56.52_{\pm 0.12}$\\
    FIAD &(ESWA $'25$) & $83.72_{\pm 0.07}$ & $87.50_{\pm 0.07}$ & $88.42_{\pm 0.09}$ & $85.19_{\pm 0.07}$ & $88.95_{\pm 0.03}$ & $77.25_{\pm 0.13}$ \\
    AHFAN &(NN $'25$) & $82.22_{\pm 0.04}$ & $87.43_{\pm 0.06}$ & $83.90_{\pm 0.05}$ & $84.17_{\pm 0.06}$ & $85.30_{\pm 0.03}$  & $79.90_{\pm 0.08}$\\
    \textbf{TERGAD} && \textcolor{red}{{$\mathbf{93.06_{\pm 0.04}}$}} & \textcolor{red}{{$\mathbf{95.78_{\pm 0.14}}$}} & \textcolor{red}{{$\mathbf{97.72_{\pm 0.08}}$}} & \textcolor{red}{{$\mathbf{94.97_{\pm 0.11}}$}} & \textcolor{red}{{$\mathbf{98.03_{\pm 0.03}}$}}  & \textcolor{red}{{$\mathbf{80.48_{\pm 0.03}}$}}\\
    \midrule
    \rowcolor{red!20} \multicolumn{8}{c}{\textbf{PR-AUC}} \\
    MLPAE & (MLSDA $'12$) & $28.91_{\pm 0.04}$ & $23.73_{\pm 0.06}$ & $35.24_{\pm 0.03}$ & $27.18_{\pm 0.10}$ & $24.01_{\pm 0.08}$ & $35.75_{\pm 0.02}$ \\
    GCNAE & (ICLR $'17$) & $27.80_{\pm 0.08}$ & $23.91_{\pm 0.04}$ & $35.00_{\pm 0.03}$ & $23.11_{\pm 0.07}$ & $23.50_{\pm 0.12}$ & $36.92_{\pm 0.04}$ \\
    DOMINANT & (SDM $'19$) & $29.98_{\pm 0.06}$ & $34.86_{\pm 0.10}$ & $39.64_{\pm 0.12}$ & $11.35_{\pm 0.08}$ & $10.25_{\pm 0.11}$ & $11.54_{\pm 0.04}$ \\
    AnomalyDAE & (ICASSP $'20$) & $28.91_{\pm 0.09}$ & $23.73_{\pm 0.10}$ & $35.24_{\pm 0.07}$ & $27.18_{\pm 0.12}$ & $24.01_{\pm 0.03}$ & $36.99_{\pm 0.03}$ \\
    CoLA & (TNNLS $'22$) & $17.62_{\pm 0.11}$ & $14.28_{\pm 0.12}$ & $13.34_{\pm 0.10}$ & $12.98_{\pm 0.06}$ & $12.25_{\pm 0.12}$ & $10.22_{\pm 0.06}$ \\
    GAAN & (CIKM $'20$) & $22.11_{\pm 0.04}$ & $23.86_{\pm 0.11}$ & $34.90_{\pm 0.12}$ & $5.41_{\pm 0.13}$ & $25.86_{\pm 0.08}$ & $30.24_{\pm 0.06}$ \\
    GADNR & (WSDM $'24$) & $30.97_{\pm 0.10}$ & $43.98_{\pm 0.05}$ & $35.68_{\pm 0.05}$ & $42.81_{\pm 0.14}$ & $40.04_{\pm 0.09}$ & $16.23_{\pm 0.06}$ \\
    FIAD & (ESWA $'25$) & $39.50_{\pm 0.07}$ & $28.24_{\pm 0.06}$ & $26.55_{\pm 0.10}$ & $15.36_{\pm 0.12}$ & $15.20_{\pm 0.09}$ & $16.93_{\pm 0.04}$ \\
    AHFAN & (NN $'25$) & $38.62_{\pm 0.10}$ & $38.08_{\pm 0.12}$ & $35.32_{\pm 0.05}$ & $26.15_{\pm 0.09}$ & $15.08_{\pm 0.14}$ & $33.57_{\pm 0.04}$ \\
    \textbf{TERGAD} && \textcolor{red}{$\mathbf{68.71}_{\pm 0.12}$} & \textcolor{red}{$\mathbf{60.07}_{\pm 0.13}$} & \textcolor{red}{$\mathbf{73.95}_{\pm 0.08}$} & \textcolor{red}{$\mathbf{50.68}_{\pm 0.10}$} & \textcolor{red}{$\mathbf{76.42}_{\pm 0.14}$} & \textcolor{red}{$\mathbf{38.25}_{\pm 0.08}$} \\
    \bottomrule
    \end{tabular}
    \label{tab:performance}
\end{table*}

\subsection{Baselines}
\par In our experiments, we compare our model with nine graph anomaly detection methods, which can be grouped into three categories:

\par Firstly, methods employ reconstruction error as an anomaly signal.
\begin{itemize} [leftmargin=0.5cm]
    \item \textbf{MLPAE}~\cite{sakurada2014anomaly} uses a multi-layer perceptron (MLP) as both encoder and decoder to reconstruct node attributes.
    \item \textbf{GCNAE}~\cite{kipf2017semi} replaces MLP with a GCN to jointly model graph structure and attributes.
    \item \textbf{DOMINANT}~\cite{ding2019deep} adopts GCN as the encoder but reconstructs the adjacency matrix via inner product and the attribute matrix via a reverse GCN layer.
    \item \textbf{AnomalyDAE}~\cite{fan2020anomalydae} introduces dedicated structure and attribute decoders to jointly learn topological and semantic patterns for anomaly detection.
\end{itemize}

\par Secondly, approaches leverage generative modeling or contrastive learning to capture anomalous beyond reconstruction.
\begin{itemize} [leftmargin=0.5cm]
    \item \textbf{GAAN}~\cite{chen2020generative} is the first to apply generative adversarial networks (GANs) to graph anomaly detection, using a generator-discriminator framework to identify outliers.
    \item \textbf{CoLA}~\cite{liu2021anomaly} utilizes contrastive learning to pull normal node pairs closer and push anomalous ones apart.
\end{itemize}

\par Thirdly, models enhance node representations through attention, neighborhood modeling, or explicit anomaly injection.
\begin{itemize} [leftmargin=0.5cm]
    \item \textbf{GADNR}~\cite{roy2024gad} emphasizes local neighborhood structures to better capture diverse anomaly types.
    \item \textbf{FIAD}~\cite{chen2025fiad} directly injects synthetic anomaly information into the feature matrix, enabling the model to learn fine-grained anomaly patterns from all nodes.
    \item \textbf{AHFAN}~\cite{wang2025graph} proposes a dual-branch framework that fuses semantic and attention-based representations to address class and semantic inconsistency.
\end{itemize}

\subsection{Experimental Setup} 
\label{sec:ES}

\par Following the standard evaluation protocol established in prior GAD literature~\cite{ding2019deep}, we adopt a synthetic anomaly injection strategy to ensure controlled and reproducible evaluation across all datasets. 
Specifically, for each dataset, we designate $5$\% of nodes as anomalous and perturb them via two complementary mechanisms:
\begin{itemize} [leftmargin=0.5cm]
    \item \textbf{Attribute Perturbation}: For each selected anomalous node, we randomly flip $30$\% of its attribute dimensions. 
    For bag-of-words or TF-IDF features (e.g., Cora, Pubmed), this corresponds to inverting the presence/absence or weight of selected terms, thereby creating semantic inconsistencies between the node's content and its structural role.
    \item \textbf{Structural Perturbation}: We rewire $20$\% of the edges incident to each anomalous node by disconnecting it from its original neighbors and reconnecting it to randomly selected nodes outside its local community. 
    This creates \textit{hub-periphery mismatches} or \textit{bridge anomalies}, where a node exhibits connectivity patterns inconsistent with its attribute profile.
\end{itemize}

\par The anomaly injection ratios for the six datasets after perturbation are detailed in Table~\ref{tab:datastatistics}.
The injected anomalies are designed to exhibit \textit{semantic-structural inconsistencies}, a hallmark of real-world anomalies in domains such as financial fraud~\cite{ma2023towards} and cyber intrusion detection~\cite{zhang2023federated}.

\par For all experiments, we set the hidden dimension of the dual-branch encoder to $64$ (the Blog dataset is set to $256$), the number of training epochs to $200$, the initial learning rate to $5 \times 10^{-3}$, and the dropout rate to $0.3$. 
The loss balancing parameter $\alpha$ is set to $0.8$, emphasizing structural reconstruction while preserving attribute fidelity.

\par To ensure statistical reliability and mitigate the impact of random initialization, we run each experiment with four different random seeds and report the mean and variance of the ROC-AUC scores. 
All models are implemented in PyTorch and trained on a single NVIDIA L$40$ GPU with the cuda backend.
The model code and precomputed LLM embeddings are publicly available to ensure reproducibility.

\subsection{Comparison Results}
\par To comprehensively evaluate the effectiveness of the proposed TERGAD framework, we compare it against nine state-of-the-art baselines across six real-world datasets. 
The quantitative results, measured by ROC-AUC (\%), are summarized in Table~\ref{tab:performance}.

\par The results demonstrate that TERGAD consistently outperforms all baseline methods across the six datasets, achieving $93.06$\%, $95.78$\%, $98.03$\%, $94.97$\%, and $97.72$\% on Cora, Citeseer, Pubmed, ACM, and DBLP, respectively. 
This consistent superiority across diverse domains, ranging from citation networks to co-authorship graphs, underscores the robustness and strong generalizability of our framework. 
Several key observations can be drawn from the results. 
First, traditional reconstruction-based methods, such as MLPAE and GCNAE, exhibit relatively lower performance, indicating their inherent limitations in capturing complex, non-linear anomalous patterns. 
Second, while FIAD performs competitively on certain datasets, its reliance on synthetic anomaly injection based on global statistics often overlooks localized structural irregularities, leading to inconsistent performance across different graph topologies.

\par The performance gain of TERGAD highlights the effectiveness of leveraging LLMs to capture structural semantic priors for anomaly detection. 
By translating abstract graph topology into descriptive natural language that articulates structural roles and connectivity patterns, our framework empowers LLMs to identify anomalies through a lens of semantic consistency. 
This advantage is particularly pronounced on the Pubmed and DBLP datasets, where our method achieves $98.03$\% and $97.72$\%, significantly surpassing all baselines. 
These results suggest that structural anomalies in academic networks often manifest as deviations from expected topological roles, such as nodes with abnormal community memberships, which are more effectively captured through semantic reasoning than pure numerical modeling.

\par Notably, TERGAD also achieves the leading performance on the ACM dataset ($94.97$\%), further confirming the efficacy of LLM-enhanced understanding in citation networks. 
This result demonstrates that our framework remains highly effective even when anomaly signals are subtle and deeply intertwined with both structural and attribute-based cues. 
In conclusion, this comprehensive comparison validates that structural information, when processed through the semantic reasoning of LLMs, provides a superior foundation for graph anomaly detection. 
Our approach offers a novel perspective by identifying structurally anomalous patterns that are often overlooked by traditional topological methods.

\par To provide a more comprehensive evaluation of TERGAD's performance, we report the Precision-Recall Area Under the Curve (PR-AUC) scores across all six benchmark datasets. 
PR-AUC is particularly informative for anomaly detection tasks due to the inherent class imbalance in these datasets. 
As shown in Table~\ref{tab:performance}, TERGAD achieves substantial improvements in PR-AUC across all datasets, with particularly notable gains on Pubmed ($76.42\%$), DBLP ($73.95\%$), and Cora ($68.71\%$). 
These results confirm that our framework effectively identifies anomalous nodes even in highly imbalanced scenarios.

\subsection{Ablation Study}
\subsubsection{Component Ablation}
\par To evaluate the contribution of individual components within the TERGAD framework, we conduct a systematic ablation study by comparing the full model against two key variants. 
The first variant, \textbf{TERGAD w/o Prompt}, omits the structured template-based prompting strategy, directly depriving the LLM of the contextual information that encodes the topological roles. 
This experiment simply concatenates structural information to the original node attributes $\mathbf{X}$.
The second variant, \textbf{TERGAD w/o Gate}, eliminates the gated fusion mechanism, replacing the adaptive node-wise balancing with a simple concatenation to integrate node attributes and LLM-derived semantic embeddings. 

\par The anomaly detection performance of these variants across six benchmark datasets is summarized in Table~\ref{tab:ablation}. 
The results consistently demonstrate that both components are indispensable for achieving optimal performance. 
Specifically, the significant performance degradation observed in the w/o Prompt variant, most notably on Cora ($17.07\%$) and Citeseer ($23.43\%$), underscores the critical role of structured prompting in eliciting high-quality semantic priors. 
By translating raw topology into role-based natural language, this strategy enables the LLM to capture nuanced structural features (e.g., hub or bridge nodes) that are vital for identifying semantic-structural inconsistencies.

\begin{table}[t]
    \centering
    \footnotesize
    \renewcommand{\arraystretch}{1.2}
    \caption{Results of the ablation study (ROC-AUC \%).}
    \label{tab:ablation_transposed}
    \begin{tabular}{lrrrrrr}
        \toprule
        Variant & Cora & Citeseer & DBLP & ACM & Pubmed  & Blog\\
        \midrule
        \rowcolor{red!20} \multicolumn{7}{c}{\textbf{Each Variants}} \\
        w/o Pro. & $75.99$ & $72.35$ & $77.12$ & $74.79$ & $92.51$ & $78.15$  \\
        w/o Gate   & $85.56$ & $85.59$ & $95.20$ & $88.50$ & $90.96$ &$79.17$ \\
        \midrule
        \rowcolor{green!20} \multicolumn{7}{c}{\textbf{Prompt Design}} \\
        w/o Deg. & $91.53$ & $93.25$ & $95.43$ & $88.89$ & $95.57$ & $79.68$ \\
        w/o T \& C & $92.65$ & $91.90$ & $95.32$ & $93.25$ & $95.16$  & $79.53$\\
        w/o $k$-C. & $86.87$ & $89.11$ & $94.20$ & $91.67$ & $94.13$  & $80.26$\\
        w/o Cen. & $89.64$ & $89.84$ & $93.90$ & $87.19$ & $94.92$  & $79.21$ \\
        w/o T-H & $91.75$ & $92.32$ & $96.21$ & $92.10$ & $97.72$ & $80.10$ \\
        \midrule
        TERGAD     & $93.06$ & $95.78$ & $97.72$ & $94.97$ & $98.03$ & $80.48$ \\
        \bottomrule
    \end{tabular}
    \label{tab:ablation}
\end{table}

\par Similarly, the inferior results of the TERGAD w/o Gate variant highlight the necessity of an adaptive fusion mechanism. 
The performance gaps are particularly pronounced on Cora ($7.50\%$) and ACM ($6.47\%$), suggesting that these datasets require a fine-grained balance between raw attributes and semantic insights. 
In the absence of the gating mechanism, the model struggles to reconcile potentially conflicting signals from the dual branches, leading to suboptimal representations that fail to exploit the synergy between structural roles and attribute information. 
Notably, the full TERGAD model achieves the best performance across all datasets, confirming that the integration of semantic enrichment and adaptive fusion is essential for robust graph anomaly detection.

\subsubsection{Impact of High-Order Structural Descriptions in Prompt Design} 
\label{sec:IHS}
\par To evaluate the contribution of individual high-order structural components within our prompting strategy, we conduct a fine-grained ablation study. 
Specifically, we systematically remove distinct topological features, including degree (Deg.), triangles and clustering coefficient (T \& C), $k$-core ($k$-C.), centrality measures (Cen.), and top-hub status (T-H), from the natural language descriptions while retaining the full TERGAD architecture. 
This allows us to isolate the semantic value provided by each structural property. 
The performance comparison, measured by ROC-AUC (\%), is presented in Table~\ref{tab:ablation}.

\par The results in Table~\ref{tab:ablation} demonstrate that every structural component contributes positively to the overall detection performance. 
The full TERGAD model consistently achieves the highest scores across all six datasets, validating the necessity of comprehensive structural semantics. 
Notably, removing the $k$-core decomposition leads to the most significant performance degradation, particularly on Cora (a drop of $6.19\%$) and Citeseer (a drop of $6.67\%$). 
This suggests that global hierarchical structural information is critical for the LLM to understand a node's coreness and robustness within the network. 
Similarly, excluding centrality measures results in substantial declines on ACM ($7.78\%$) and Cora ($3.42\%$), indicating that metrics quantifying node influence (e.g., betweenness, closeness) are vital for identifying anomalies that deviate from expected influence patterns.

\par Furthermore, the removal of local cohesion metrics (Triangles \& Clustering Coefficient) and connectivity basics (Degree) also yields consistent performance drops across all datasets, though to a varying extent. 
For instance, on ACM, removing degree information causes a $6.08\%$ decrease, highlighting that even basic connectivity counts are foundational for semantic grounding. 
The relatively smaller when removing Top-Hub status imply that while explicit hub ranking is beneficial, its information is partially correlated with degree and centrality. 
Overall, these findings confirm that enriching prompts with diverse, multi-scale structural descriptors, from local connectivity to global hierarchy, enables the LLM to construct more nuanced semantic representations, thereby enhancing the robustness of anomaly detection.

\begin{table}[t]
    \centering
    \footnotesize
    \renewcommand{\arraystretch}{1.2}
    \caption{Performance comparison of TERGAD with different module variants (ROC-AUC \%).}
    \label{tab:module variants}
    \begin{tabular}{lcccccc}
    \toprule
    Method & Cora & Citeseer & DBLP & Pubmed &ACM  & Blog \\
    \midrule
    \rowcolor{red!20} \multicolumn{7}{c}{\textbf{Fusion Strategies}} \\
    Concat         & $85.56$ & $85.59$ & $95.20$ & $88.50$ & $90.96$ & $79.17$ \\
    Add            & $84.44$ & $83.15$ & $90.40$ & $81.22$ & $95.60$ & $78.57$ \\
    Multiply       & $80.09$ & $79.70$ & $81.45$ & $90.30$ & $95.44$ & $78.34$ \\
    Attention      & $89.26$ & $91.26$ & $93.46$ & $86.21$ & $82.18$  & $78.89$\\
    T-A & $89.77$ & $91.84$ & $93.95$ & $91.56$ & $90.08$  & $78.93$\\
    \midrule    
    \rowcolor{green!20} \multicolumn{7}{c}{\textbf{LLM}} \\
    Qwen$3$-$4$B & $83.29$ & $71.55$ & $92.43$ & $87.54$ & $95.43$ & $79.62$\\
    nomic-v$1.5$ & $89.60$ & $91.19$ & $96.94$ & $91.88$ & $91.12$  &$79.23$\\
    \midrule
    \rowcolor{blue!20} \multicolumn{7}{c}{\textbf{Prompt Order}} \\
    shuffled & $91.92$ & $89.69$ & $96.18$ & $91.99$ & $94.65$ & $80.34$\\
    structured & $85.56$ & $93.52$ & $94.78$ & $92.17$ & $97.96$  &$80.18$\\
    \midrule
    TERGAD & $93.06$ & $95.78$ & $97.72$ & $94.97$ & $98.03$ & $80.48$ \\
    \bottomrule
    \end{tabular}
\end{table}

\subsubsection{Effect of Different Module Variants}
\par To investigate the impact of different fusion strategies in our dual-branch architecture, we conduct a comprehensive study comparing five representative fusion mechanisms: 
(1) \textbf{Concatenation} (Concat), which horizontally concatenates the original node attributes and the LLM-derived semantic embeddings; 
(2) \textbf{Element-wise Addition} (Add), which performs dimension-aligned addition of the two feature matrices; 
(3) \textbf{Element-wise Multiplication} (Multiply), which computes the Hadamard product of the two feature matrices; 
(4) \textbf{Attention mechanism} (Attention), which uses a standard attention mechanism to compute adaptive weights;
and (5) \textbf{Temperature-scaled Attention} (T-A), which introduces a learnable temperature parameter $\tau$ to sharpen or smooth the attention distribution, i.e., $\text{softmax}(\mathbf{q}^\top \mathbf{k} / \tau)$.

\par For the Add and Multiply operations, when the dimensions of original attributes and LLM embeddings differ, our implementation automatically applies PCA dimensionality reduction to align both representations to the minimum dimension before fusion. 
All fusion variants operate on LLM embeddings that have been standardized using Z-score normalization to ensure stable training. 
All variants share the same overall architecture and training protocol, differing only in the fusion module. 
The results, measured by ROC-AUC (\%), are summarized in Table~\ref{tab:module variants}.

\par Comparing the results of our fusion strategies with the baseline methods in Table~\ref{tab:performance}, we observe consistent and significant improvements. 
For instance, on the Cora dataset, our Temp-Attention fusion achieves $89.77$\%, outperforming the best baseline DOMINANT ($88.52$\%) by $1.25$ points. 
Similarly, on Pubmed, both Add ($95.60$\%) and Multiply ($95.44$\%) fusion strategies exceed the best baseline GAAN ($90.35$\%) by approximately $5$ points. 
Even on datasets like ACM where the baseline performance is relatively strong, our Temp-Attention method ($91.56$\%) still surpasses the best baseline GAAN ($88.25$\%) by $3.31$ points. 
These results highlight that integrating LLM-derived semantic embeddings through adaptive fusion mechanisms consistently enhances anomaly detection performance beyond state-of-the-art baselines.

\par To assess how the choice of LLMs affects performance, we followed the identical experimental pipeline with two alternative semantic encoders: \textbf{Qwen$3$-Embedding-$4$B} and \textbf{nomic-embed-text-v$1.5$}. 
The only difference in this study lies in the model used to generate the initial semantic node embeddings from textual node descriptions. 
The rest of the framework architecture and training procedure, comprising gated fusion, dual-branch autoencoding, and anomaly score calculation, was kept strictly consistent.
The comparative results are presented in Table~\ref{tab:module variants}.

\par The results, as presented in Table~\ref{tab:module variants}, demonstrate that both Qwen$3$-Embedding-$4$B and nomic-embed-text-v$1.5$ enable our TERGAD framework to achieve strong performance, outperforming non-LLM baseline methods on most datasets (as shown in Table~\ref{tab:performance}). 
Notably, Qwen$3$-Embedding-$4$B achieves superior performance on two datasets (Citeseer and Pubmed), while nomic-embed-text-v$1.5$ performs best on the other three (Cora, ACM, and DBLP). 
Across all datasets, nomic-embed-text-v$1.5$ attains a marginally higher average ROC-AUC. 
This indicates that both LLMs are capable semantic encoders, with each showing particular strengths on different types of graph data. 
The competitive performance of nomic-embed-text-v$1.5$, despite its significantly smaller parameter size ($0.65$B vs. $4$B), suggests that model scale is not the sole determinant of effectiveness for this task.

\par It is worth noting that, although these two LLM variants achieve competitive results, they do not outperform our primary model employing BGE-large-en-v$1.5$ embeddings, corresponding to the last row (TERGAD) in Table~\ref{tab:module variants}, which attains the best results across all datasets (a mean AUC of $93.06$\% on Cora, $95.78$\% on Citeseer, $98.03$\% on Pubmed, $94.97$\% on ACM, and $97.72$\% on DBLP). 
This suggests that BGE, despite being a smaller model, is particularly well-suited for capturing the semantic signals relevant to graph anomaly detection in our setup. 
In summary, the choice of LLM remains a critical factor, and our results highlight the importance of selecting an LLM whose pretraining domain aligns with the target application.

\subsubsection{Prompt Analysis}
\par Our node descriptions are composed of multiple structural sentences. 
To assess whether the~\textbf{order} in which these sentences appear affects model performance, we evaluate TERGAD under a shuffled prompt configuration, where the sequence of descriptive clauses is randomly permuted for each node while preserving all content.

\par
\begin{figure}[h]
    \centering
    \subfigure{
		\begin{minipage}[b]{0.45\textwidth}
			\centering
			\includegraphics[width=1\textwidth]{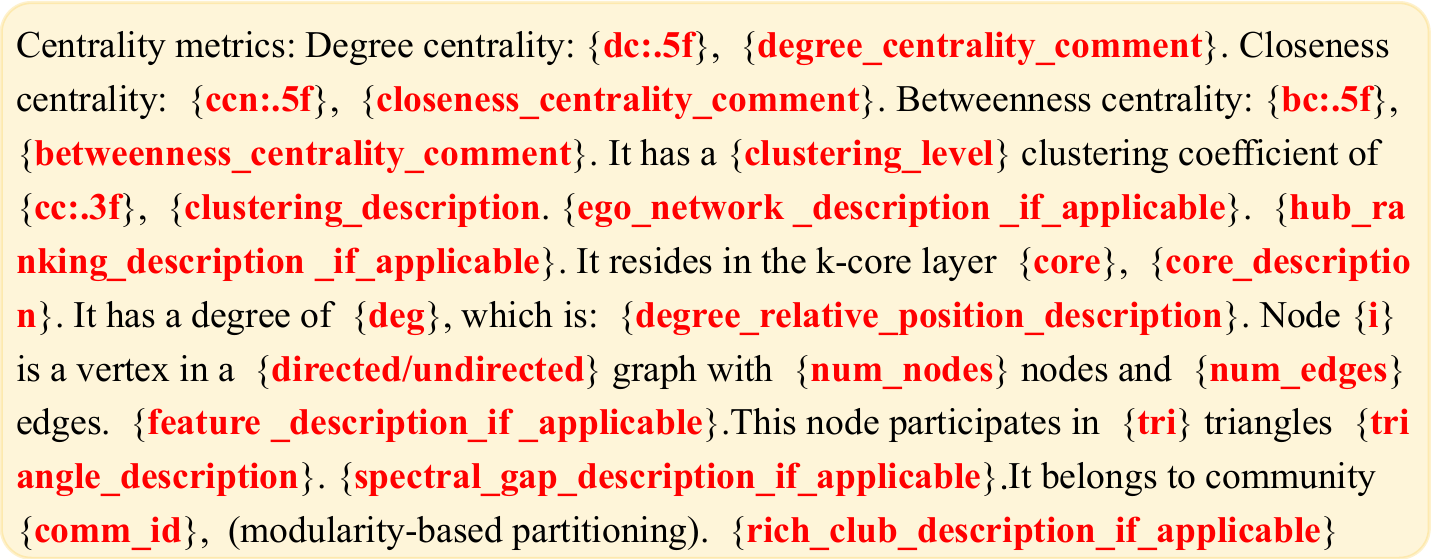}
    	\end{minipage}
	}
	\subfigure{
		\begin{minipage}[b]{0.45\textwidth}
			\centering
			\includegraphics[width=1\textwidth]{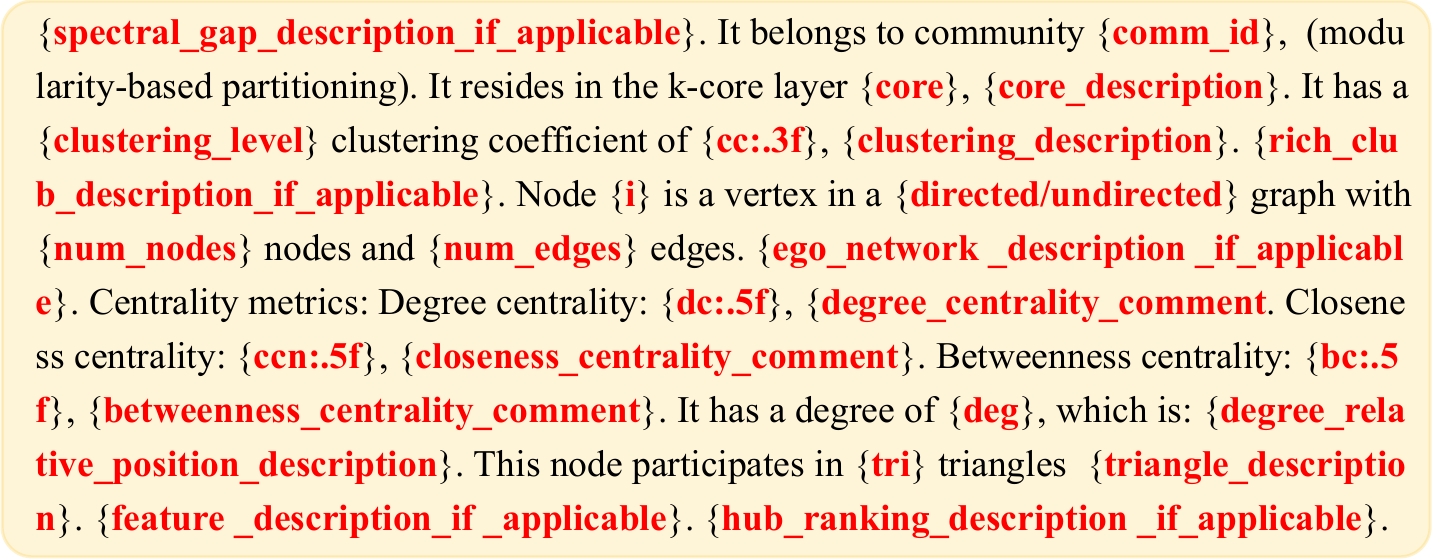}
		\end{minipage}
	}
    \subfigure{
		\begin{minipage}[b]{0.45\textwidth}
			\centering
			\includegraphics[width=1\linewidth]{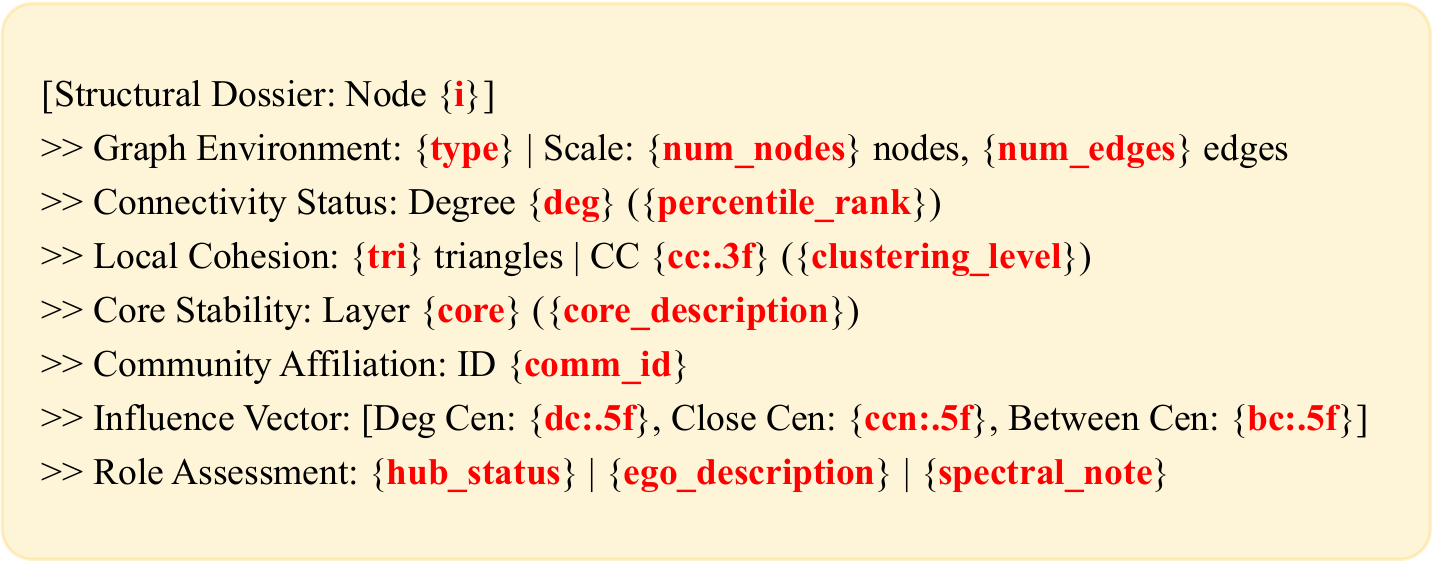}
		\end{minipage}
	}
    \caption{Illustration of Node-level Prompt Order Divergence. 
    Two random realizations of per-node prompt shuffling, demonstrating how independent randomization leads to distinct input sequences for different nodes.}
    \label{fig:prompt}
\end{figure}

 Fig.~\ref{fig:prompt} shows two independent trials of this per-node shuffling process. 
These are representative samples among many random configurations tested. 
As summarized in Table~\ref{tab:module variants}, TERGAD achieves strong performance under shuffled prompts.
This is only slightly lower than the results obtained with the original structured prompt. 

\par These findings confirm that TERGAD’s performance is largely invariant to the syntactic ordering of structural facts in the prompt, suggesting that the LLM effectively aggregates semantic signals regardless of surface-level sentence arrangement. 
This robustness further supports the reliability of our text-enhanced representation strategy.

\par To investigate how different natural language formulations affect the quality of LLM-derived structural semantics, we conduct a comparative study between our original narrative-style prompt and an alternative structured profile-style template. 
The profile-style template adopts a technical dossier format with explicit section headers and delimiter-based organization (as shown in Fig.~\ref{fig:prompt}), contrasting with the original fluent paragraph-style descriptions. 
This experiment isolates the impact of linguistic presentation on the downstream anomaly detection performance, while keeping all other components identical.

\par As shown in Table~\ref{tab:module variants}, the original narrative-style prompt consistently outperforms the structured profile-style template across all six datasets, with an average improvement of $4.53\%$. 
The performance gap is most pronounced on Cora ($7.50\%$) and DBLP ($2.94\%$), suggesting that fluent, self-contained natural language descriptions better align with the pretraining distribution of off-the-shelf LLMs. 
We hypothesize that narrative-style prompts, which mimic human-readable explanatory text, enable LLMs to more effectively leverage their inherent linguistic reasoning capabilities for structural understanding. 
In contrast, the profile-style template, while more compact and machine-readable, may introduce syntactic patterns that deviate from the LLM's pretraining corpus, slightly impairing semantic extraction quality.

\par Notably, on Pubmed, both prompt styles achieve nearly identical performance ($98.03\%$ vs. $97.96\%$), indicating that for larger graphs with rich attribute information, the specific linguistic formulation becomes less critical. This observation suggests that the semantic signal from node attributes may dominate the structural signal in such scenarios, reducing the sensitivity to prompt design. Overall, these findings validate our design choice of using narrative-style prompts and highlight the importance of aligning prompt formulations with LLM pretraining characteristics for optimal structural semantic extraction.

\subsection{Hyperparameter Analysis}
\par We also investigate the impact of model depth by comparing single-layer and two-layer GCN variants for both the encoder and decoder. 
The experimental results demonstrate that the single-layer configuration consistently outperforms its two-layer counterpart across all six evaluated datasets. 

\par To ensure a fair and controlled comparison, the two-layer GCN variant strictly adheres to the experimental setup delineated in Section~\ref{sec:ES}. 
Specifically, we extend the original single-layer architecture by sequentially stacking an additional GCN layer for each branch (i.e., the attribute branch and the LLM embedding branch), while maintaining identical hyperparameter configurations (learning rate, hidden dimension, dropout rate) and training protocols. 
The two-layer configuration adopts a standard stacked GCN architecture (Layer $1$ $\rightarrow$ ReLU $\rightarrow$ Layer $2$ $\rightarrow$ ReLU) without incorporating residual connections or skip connections. 
This methodological choice ensures that any observed performance discrepancies can be unequivocally attributed to variations in model depth rather than confounding factors such as architectural modifications or hyperparameter heterogeneity.

\par
\begin{figure}[h]
    \centering
    \includegraphics[width=0.9\linewidth]{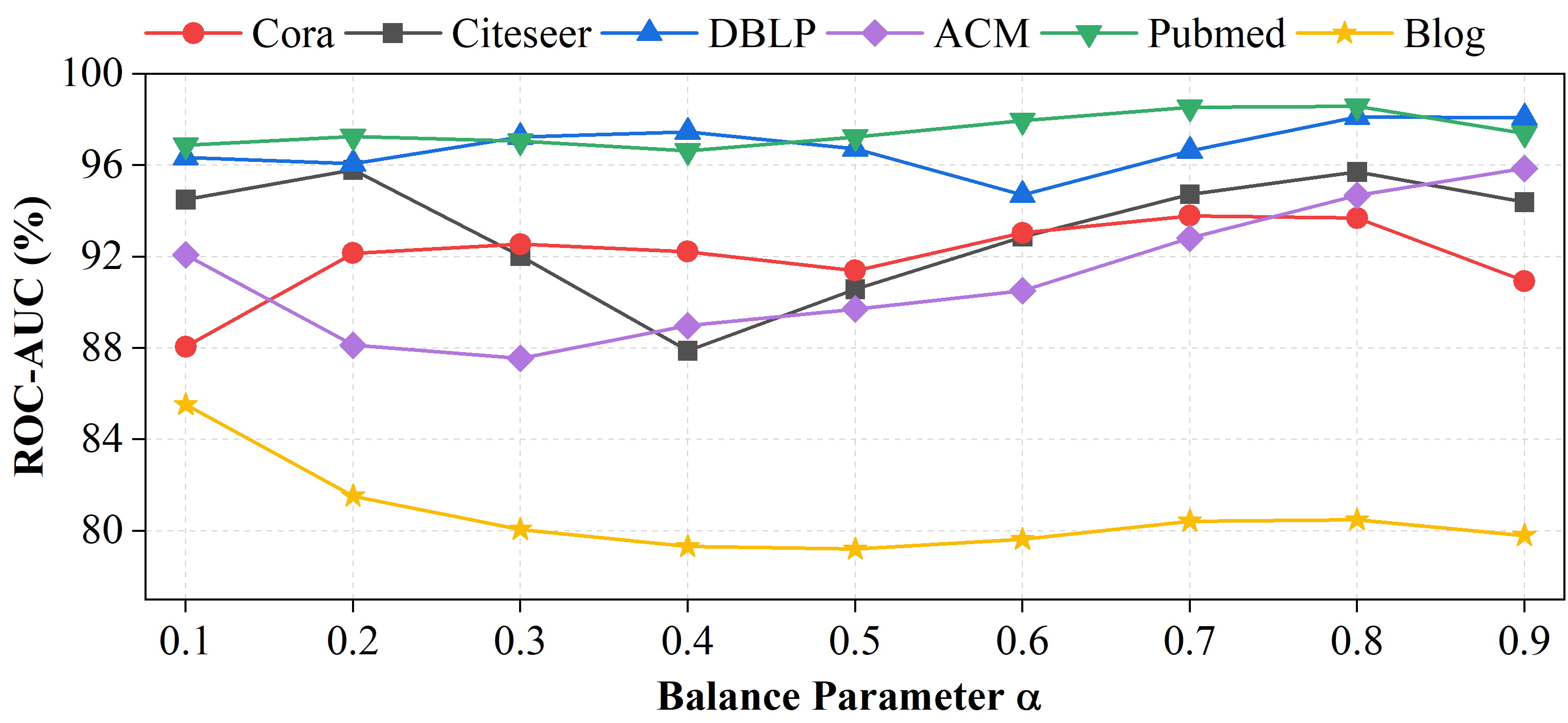}
    \includegraphics[width=0.9\linewidth]{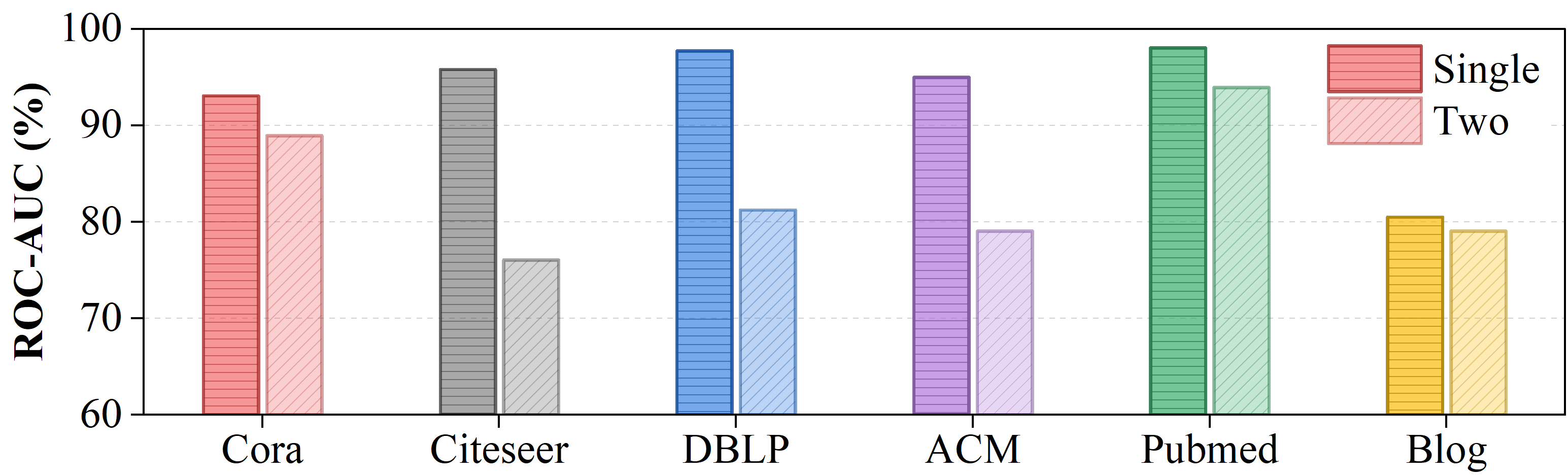}
    \caption{Performance comparison of TERGAD across six datasets under varying balance parameter $\alpha$, and varying layer GCN variants.}
    \label{fig:hyperparamter} 
\end{figure}

 We attribute this performance degradation to the increased model complexity introduced by the additional layer, which may lead to over-smoothing or information redundancy, particularly when the LLM-derived semantic embeddings already provide rich structural context. 
This observation, supported by the empirical evidence in Fig.~\ref{fig:hyperparamter}, further validates our choice of a shallow, single-layer architecture as the optimal balance between expressiveness and sensitivity to anomalous patterns.

\par Furthermore, we conduct a study to evaluate the sensitivity of our model to the balance parameter $\alpha$. 
As illustrated in Fig.~\ref{fig:hyperparamter}, the ROC-AUC scores remain consistently high across a wide range of $\alpha$ values (from $0.1$ to $0.9$) for all six datasets. 
Notably, the performance peaks around $\alpha = 0.8$, but remains robust even when $\alpha$ varies between $0.6$ and $0.9$, with minimal degradation. 
This demonstrates that our LLMforGAD framework is not overly sensitive to the exact choice of $\alpha$, indicating its stability and generalizability. 
The consistent top performance at $\alpha=0.8$ further validates our default setting, while the flat performance curve in the neighborhood of this value suggests that our method can maintain strong detection accuracy even under suboptimal parameter configurations.

\subsection{Efficiency Analysis}
\subsubsection{Generation Time Consumption Analysis}
\label{sec:TCA}
\par To evaluate the computational efficiency of our text generation and embedding process, we measure the time required for two key steps on six benchmark datasets. 
The first step is the generation of structured, natural language descriptions for each node based on its graph structural properties. 
The second step is the computation of embeddings from these generated texts using the BGE model.

\par As shown in Table \ref{tab:training_time}, the text generation phase is efficient, taking less than $0.25$ seconds even for the largest dataset (Pubmed, with $19,717$ nodes). 
This demonstrates that our method for converting complex graph topology into descriptive text is highly scalable and incurs minimal overhead. 
In contrast, the embedding computation is the primary time cost, which is expected as it involves a large pre-trained language model. 
The time for this step scales linearly with the number of nodes, confirming the predictable computational demand of our approach. 

\subsubsection{Training Time Consumption Analysis}
\par We measure the total training time (in seconds) for $200$ epochs across all methods and datasets. 
This metric reflects the computational efficiency of each approach during the training phase. 
The results are presented in Table~\ref{tab:training_time}.

\begin{table}[h]
\centering
\footnotesize
    \renewcommand{\arraystretch}{1.2}
    \caption{Training time (seconds) for $200$ epochs across six datasets.}
    \label{tab:training_time}
    \setlength{\tabcolsep}{3pt}
    \begin{tabular}{lrrrrrr}
        \toprule
        Method & Cora & Citeseer & DBLP & ACM & Pubmed & Blog \\
        \midrule
        \rowcolor{red!20} \multicolumn{7}{c}{\textbf{Pre-processing}} \\
        Generation & $0.07$ & $0.14$ & $0.10$ & $0.19$ & $0.21$ & $0.15$ \\
        Embedding & $20.44$ & $24.62$ & $40.52$ & $117.07$ & $140.58$ & $46.30$ \\
        \midrule
        \rowcolor{green!20} \multicolumn{7}{c}{\textbf{In-processing}} \\
        MLPAE  & $3.11$ & $4.33$ & $6.50$ & $22.62$ & $5.62$ & $7.71$\\
        GCNAE  & $5.34$ & $6.80$ & $12.68$ & $8.25$ & $8.20$ & $30.32$ \\
        DOMINANT  & $6.46$ & $8.92$ & $16.02$ & $2,361.97$ & $331.30$ & $30.65$ \\
        AnomalyDAE  & $5.47$ & $6.05$ & $10.44$ & $505.47$ & $811.85$ & $13.17$\\
        CoLA & $4.61$ & $6.11$ & $9.36$ & $23.23$ & $7.09$  & $13.13$\\
        GAAN  & $11.75$ & $14.10$ & $174.31$ & $5,663.09$ & $435.07$ & $132.40$ \\
        GADNR  & $111.31$ & $136.40$ & $1,771.10$ & $1,071.41$ & $739.53$  & $26.13$\\
        FIAD  & $6.48$ & $19.54$ & $69.22$ & $351.49$ & $326.39$ & $49.84$\\
        AHFAN  & $4.88$ & $5.12$ & $4.16$ & $29.95$ & $20.92$ & $8.68$\\
        \midrule
        \textbf{TERGAD} & $1.77$ & $2.45$ & $7.16$ & $52.70$ & $21.98$ &$4.52$\\
        \bottomrule
    \end{tabular}
\end{table}

\par TERGAD demonstrates competitive computational efficiency, achieving the fastest training time on smaller datasets like Cora ($1.77$s) and Citeseer ($2.45$s). 
On larger datasets (DBLP, ACM, Pubmed), while lightweight autoencoders (e.g., MLPAE, GCNAE, AHFAN) exhibit lower training times due to simpler architectures, TERGAD remains significantly more efficient than complex baselines such as DOMINANT, GAAN, and GADNR. 
For instance, on ACM, TERGAD completes training in $52.70$s, substantially faster than DOMINANT ($2,361.97$s) and GAAN ($5,663.09$s). 
This efficiency stems from our streamlined dual-branch architecture, which avoids computationally expensive adversarial training or extensive neighborhood reconstruction. 
The modest overhead compared to simple autoencoders is justified by the substantial performance gains in detection accuracy, highlighting a favorable trade-off between efficiency and effectiveness.

\subsubsection{Memory Allocation Analysis}
\par We report the maximum GPU memory allocated (in MB) during training for each method. 
Additionally, we provide detailed memory consumption analysis for TERGAD's LLM embedding generation phase. 
The results are shown in Tables~\ref{tab:memory_allocation}.

\begin{table}[h]
\centering
    \renewcommand{\arraystretch}{1.2}
    \caption{Maximum GPU memory allocated (MB) during training across six datasets.}
    \label{tab:memory_allocation}
    \setlength{\tabcolsep}{4pt}
    \footnotesize
    \begin{tabular}{lrrrrrr}
        \toprule
        Method & Cora& Citeseer & DBLP & ACM& Pubmed &Blog \\
        \midrule
        MLPAE  & $94$ & $275$ & $732$ & $2,657$ & $214$ & $839$ \\
        GCNAE  & $235$ & $590$ & $1,812$ & $864$ & $613$ & $22,668$ \\
        DOMINANT  & $300$ & $695$ & $2,055$ & $14,686$ & $9,124$ & $22,847$  \\
        AnomalyDAE  & $284$ & $510$ & $1,424$ & $9,047$ & $10,653$ & $1,584$ \\
        CoLA  & $87$ & $229$ & $612$ & $2,139$ & $183$ & $723$ \\
        GAAN & $2,258$ & $2,746$ & $7,860$ & $8,558$ & $4,318$ & $10,325$\\
        GADNR  & $3,910$ & $4,810$ & $7,979$ & $7,556$ & $7,849$ &$2,123$\\
        FIAD& $438$ & $874$ & $2,463$ & $15,170$ & $15,251$ & $2,573$\\
        AHFAN & $61$ & $94$ & $264$ & $1,104$ & $265$  & $1,275$ \\
        \textbf{TERGAD} & $277$ & $494$ & $1,374$ & $8,960$ & $10,643$ &$1,523$ \\
        \bottomrule
    \end{tabular}
\end{table}

\par Table~\ref{tab:memory_allocation} presents a comprehensive comparison of GPU memory consumption across all methods. 
We observe that lightweight autoencoder-based approaches (e.g., AHFAN, MLPAE, CoLA) exhibit lower memory footprints due to their simpler architectures and absence of semantic enrichment modules. 
However, these methods achieve such efficiency at the cost of limited representational capacity, as evidenced by their comparatively lower detection performance in Table~\ref{tab:performance}.

\par In contrast, TERGAD strikes a favorable balance between memory efficiency and detection effectiveness. 
On smaller datasets (Cora: $276.53$ MB, Citeseer: $494.21$ MB, DBLP: $1,374.29$ MB), TERGAD maintains memory consumption within a practical range while delivering substantial performance gains. 
Even on larger datasets like ACM and PubMed, where memory usage increases to $8,960.41$ MB and $10,643.45$ MB, respectively. 
TERGAD remains significantly more memory-efficient than comparably powerful baselines such as FIAD ($15,170.38$ MB and $15,250.86$ MB), DOMINANT ($14,686.25$ MB and $9,123.77$ MB), and GAAN ($8,558.05$ MB and $4,318.37$ MB).

\par The base BGE-large-en-v$1.5$ model requires $1,257$ MB for initialization. 
During batch processing ($100$ nodes per batch), the peak memory consumption remains stable across all datasets (approximately $2,592$--$2,598$ MB), demonstrating that our embedding generation process is memory-efficient and scalable regardless of graph size. 
This one-time preprocessing cost is acceptable given the substantial performance gains achieved by incorporating LLM-derived semantic embeddings.

\par In conclusion, the primary objective of Fig. 1 is to address a fundamental question: \textit{Can LLMs alone, without explicit graph structure modeling, effectively detect graph anomalies?} 
To investigate this, we conducted a controlled experiment on the standard text-rich graph dataset Citeseer.
\par We compare the following methods:
\begin{itemize} [leftmargin=0.5cm]
    \item DOMINANT~\cite{ding2019deep}, a strong GCN-based graph anomaly detection method achieving AUC = $92.85$\% (see Table~\ref{tab:performance}).
    \item ChatGPT-$4.1$, Gemini-$3$, Deepseek, and Qwen3-Max, all operating in a zero-shot setting. Specifically, we provide each LLM with raw node attributes (bag-of-words) and the complete adjacency matrix in textual format (e.g., ``node $i$ connects to [$j_1, j_2, \dots$]''), and request binary anomaly labels for each node.
    \item We report Recall@K, where $K$ equals the number of ground-truth anomalous nodes ($K = 150$).
\end{itemize}

\par As shown in Fig. 1, all LLM-based methods exhibit Recall@K scores that are more than 70\% lower than DOMINANT. This substantial performance gap confirms our hypothesis: \textit{LLMs alone cannot reliably detect graph anomalies, even on text-rich graphs, due to their lack of explicit graph reasoning mechanisms.} 

\par This finding directly motivates the core design principle of TERGAD: rather than serving as an independent detector, the LLM functions as a semantic augmentation module that generates high-quality embeddings from structured textual descriptions, which are then adaptively fused with structural information through our dual-branch gated autoencoder.

\section{Conclusion}
\par In this paper, we propose TERGAD, a novel framework for graph anomaly detection that bridges the gap between graph topology and semantic reasoning through LLMs. 
Grounded in the insight that natural language can effectively encapsulate high-level structural roles, such as ``hubs", ``bridges", or ``peripheral nodes", we first transform abstract graph structures into descriptive textual narratives, which are subsequently encoded by an LLM into rich semantic embeddings for enhanced data representation. 
To synergize these insights with raw node attribute data, we introduce a gated dual-branch autoencoder that adaptively fuses both modalities to reconstruct the original attributes and graph structure. 
The resulting anomaly scores are derived from the joint reconstruction error, ensuring that the detection process is both interpretable and aligned with the fundamental definitions of graph anomalies. 
Notably, TERGAD is a flexible, plug-and-play framework that requires no LLM fine-tuning and supports efficient end-to-end training.
Extensive experiments on six real-world graph datasets demonstrate that TERGAD consistently outperforms baselines. 
In the future, we will explore dynamic prompt optimization, multi-hop semantic propagation, and the extension of this framework to heterogeneous and temporal graph settings for broader data mining applications.

\bibliography{LLMforGAD}
\bibliographystyle{IEEEtran}


\vspace{-12pt}
\begin{IEEEbiography}[{\includegraphics[width=1in,height=1.25in,clip,keepaspectratio]{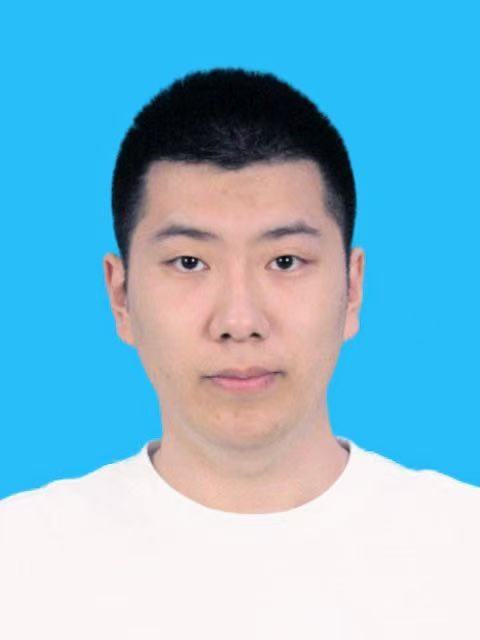}}] {Wen Shi} is currently a Master student in College of Software Engineering, Jilin University, Changchun, China.
Before that, he received the B.Sc. degree from Northeast Agricultural University, Harbin, China, in 2024.
His research interests include graph learning and large language models.
\end{IEEEbiography}

\vspace{-12pt}
\begin{IEEEbiography}[{\includegraphics[width=1in,height=1.25in,clip,keepaspectratio]{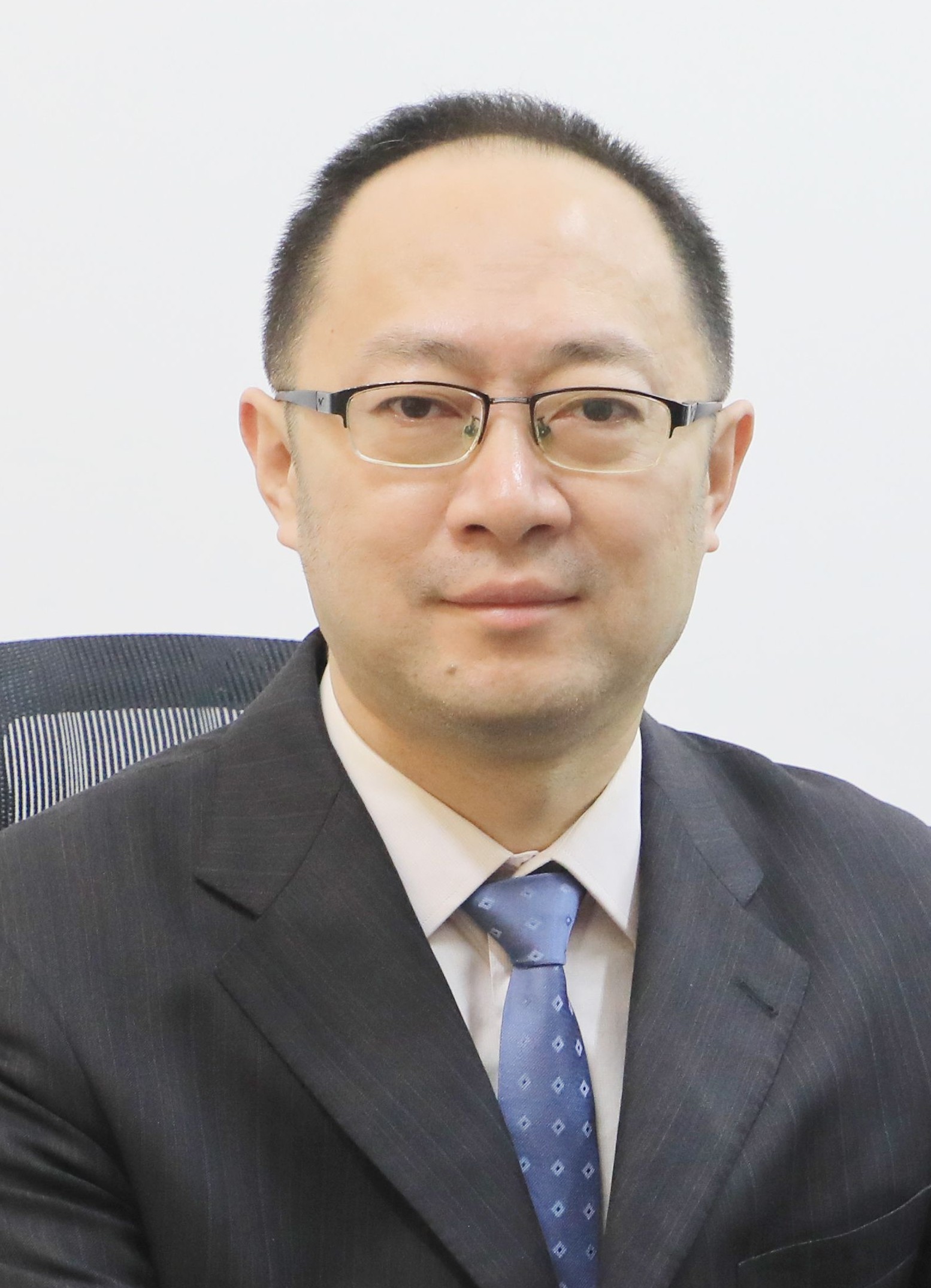}}] {Zhe Wang} received the B.Sc., M.Sc. and Ph.D. degrees from Jilin University, Changchun, China, in 1997, 2001 and 2005, respectively. 
He is currently a Full Professor with the College of Computer Science and Technology, Jilin University, Changchun, China.
He has published over 50 scientific papers in international journals and confernce.
His research interests include artificial intelligence, data mining, social media mining, business intelligence.
\end{IEEEbiography}

\vspace{-12pt}
\begin{IEEEbiography}[{\includegraphics[width=1in,height=1.25in,clip,keepaspectratio]{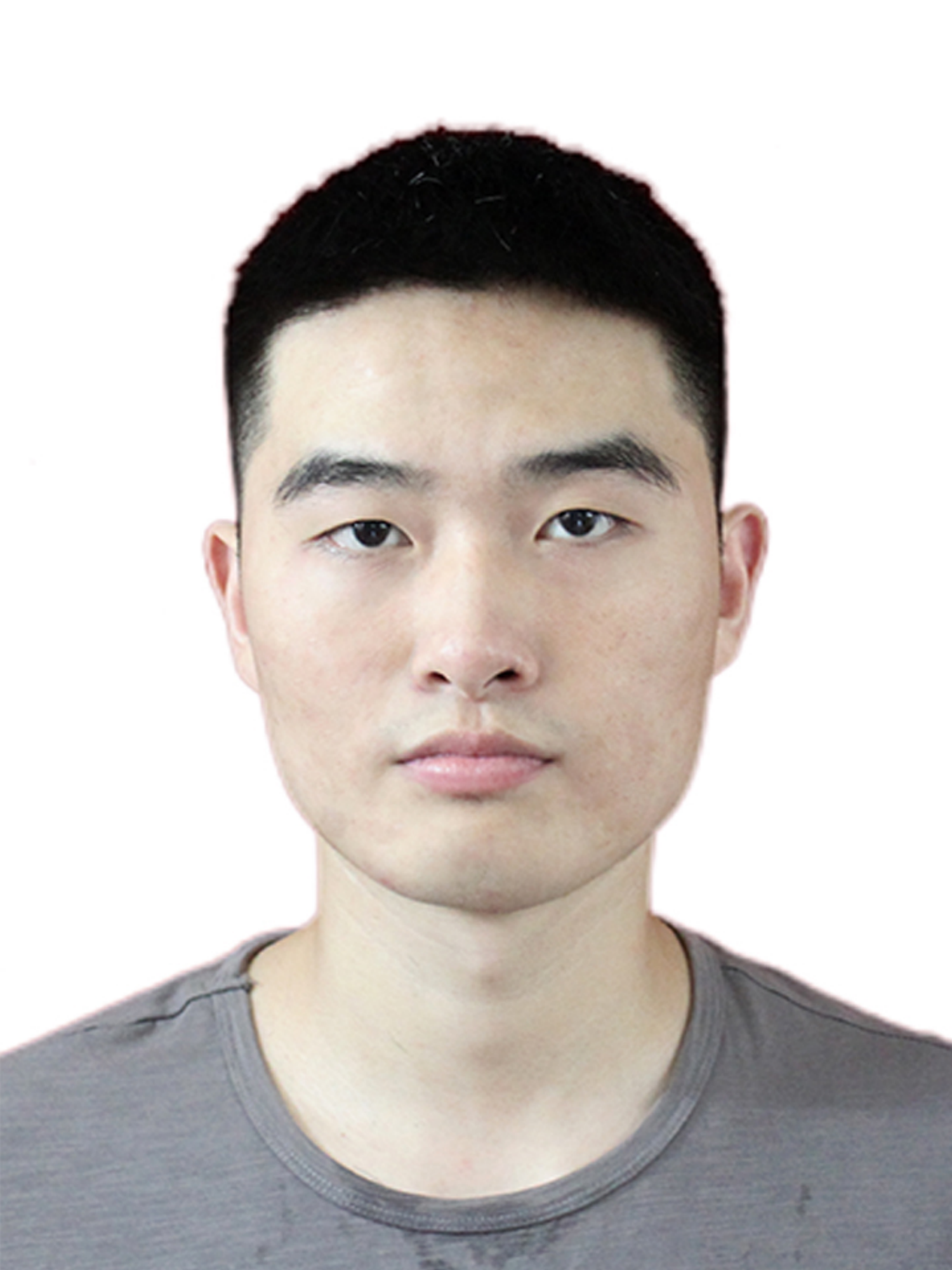}}]{Huafei Huang}
    (Graduate Student Member, IEEE) received the BSc degree in Internet of Things engineering from the North University of China in 2020, and the MSc degree in software engineering from the Dalian University of Technology in 2023. He is currently working toward the PhD degree with the School of Computer Science and Information Technology, Adelaide University. His research interests  include graph learning and large language models.
\end{IEEEbiography}

\vspace{-12pt}
\begin{IEEEbiography}[{\includegraphics[width=1in,height=1.25in,clip,keepaspectratio]{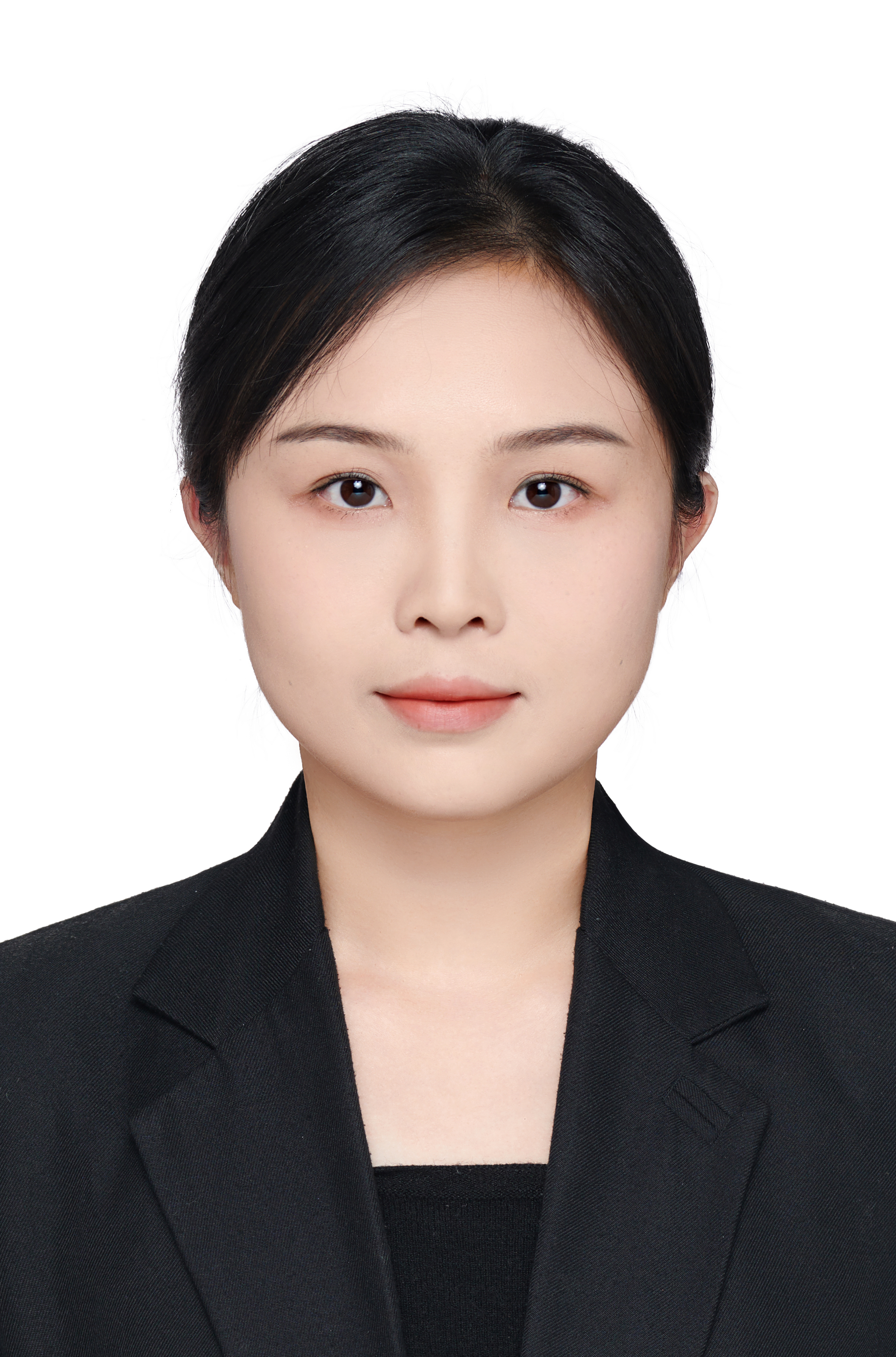}}] {Qing Qing} is currently a PhD student in College of Computer Science and Technology, Jilin University, Changchun, China.
Before that, she received the B.Sc. degree from Northeast Agricultural University, Harbin, China, in 2018, and the M.Sc. degree from Dalian University of Technology, Dalian, China, in 2021. 
Her research interests include graph learning, algorithmic fairness, responsible AI.
\end{IEEEbiography}

\vspace{-12pt}
\begin{IEEEbiography}[{\includegraphics[width=1in,height=1.25in,clip,keepaspectratio]{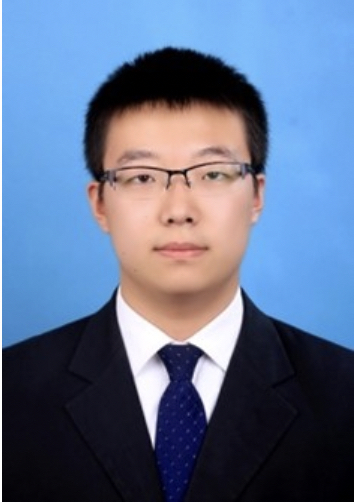}}]{Ziqi Xu} received the M.S. degree in Computing and Innovation from the School of Computer and Mathematical Sciences, The University of Adelaide, Australia, and the Ph.D. degree in Computer Science from the University of South Australia, Australia. 
He is currently a Lecturer in Data Science and Artificial Intelligence with the School of Computing Technologies, RMIT University, Australia. 
His research interests include responsible AI, causal inference, fairness, and explainable machine learning.
\end{IEEEbiography}

\vspace{-12pt}
\begin{IEEEbiography}[{\includegraphics[width=1in,height=1.25in,clip,keepaspectratio]{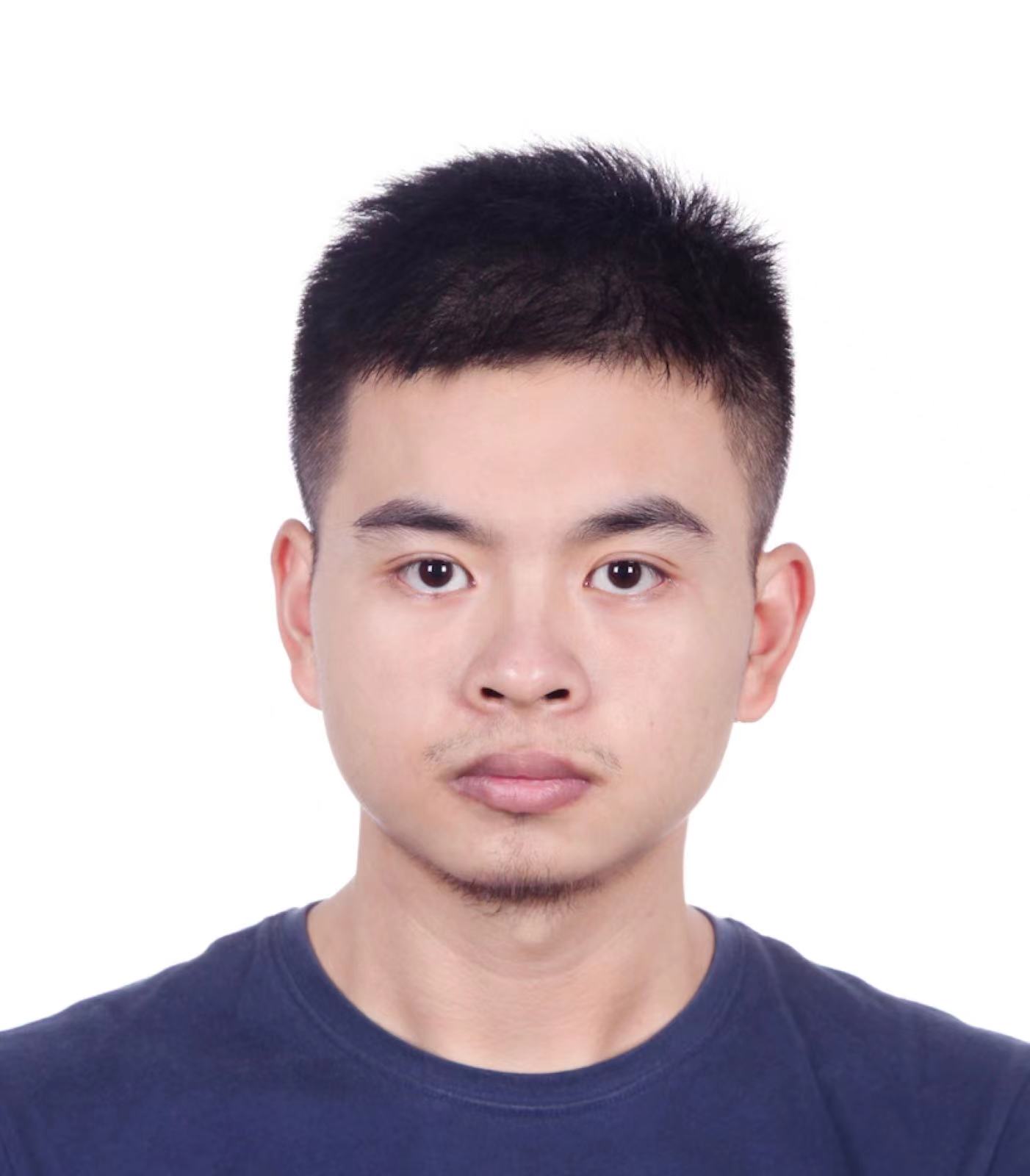}}]{Qixin Zhang }
received his B.S. degree from the University of Science and Technology of China in 2018. He subsequently earned his Ph.D. degree in the College of Computing at City University of Hong Kong in 2024. Currently, he is a Research Fellow at Nanyang Technological University, Singapore. His research interests include optimization, subset selection, online learning and large language models. He has published over 25 papers in top-tier venues such as ICML, NeurIPS, ICLR, CVPR, ACL and TKDE.
\end{IEEEbiography}

\vspace{-12pt}
\begin{IEEEbiography}[{\includegraphics[width=1in,height=1.25in,clip,keepaspectratio]{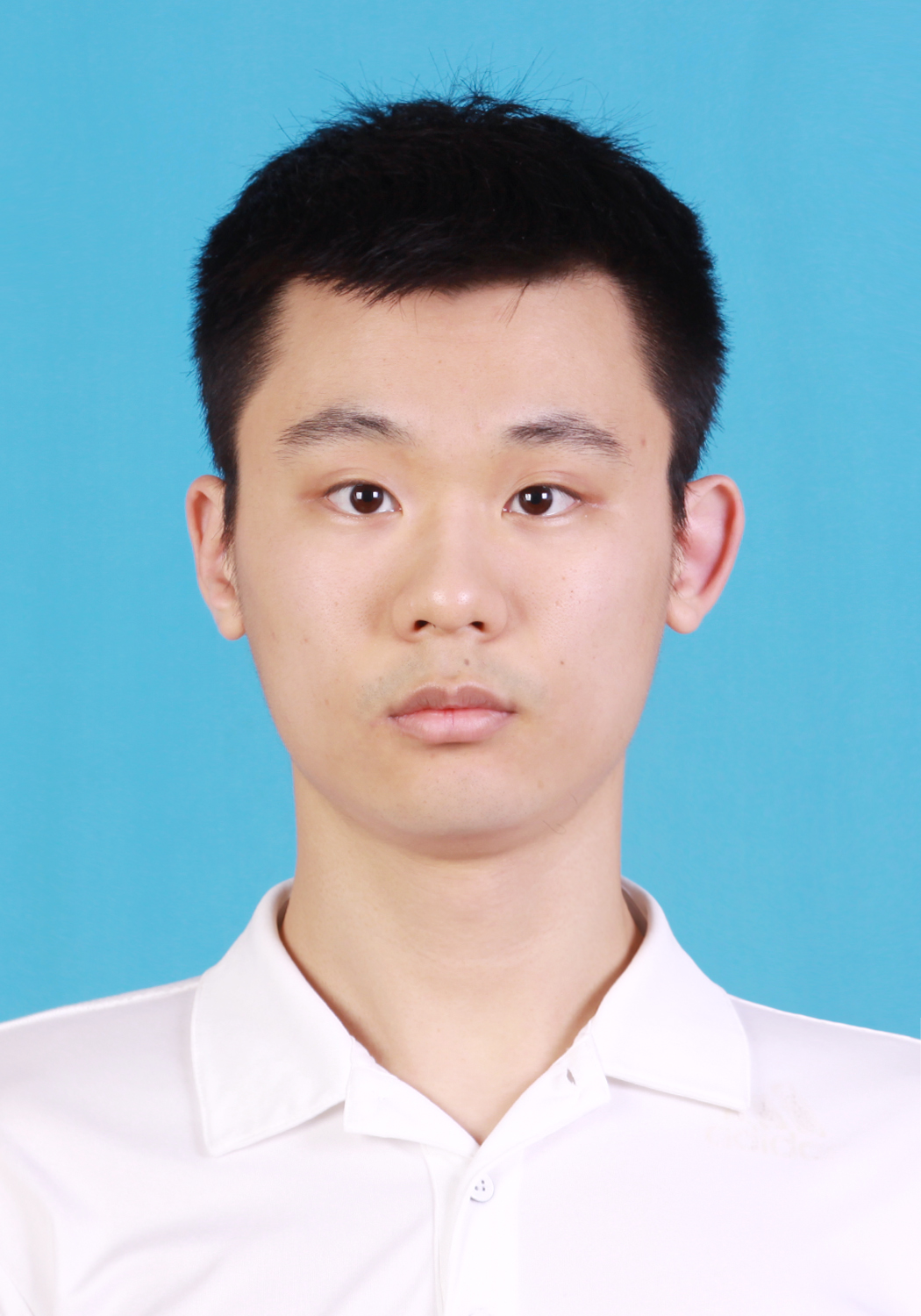}}]{Xikun Zhang} is a Lecturer at the School of Computing Technologies at RMIT University. He received his Ph.D. from the School of Computer Science at the University of Sydney. His research interests span deep graph learning, reasoning with large language models, and biomedical AI. His work has been published in leading conferences and journals, including ICLR, NeurIPS, KDD, ICDM, CVPR, ECCV, TPAMI, and TNNLS.
\end{IEEEbiography}

\vspace{-12pt}
\begin{IEEEbiography}[{\includegraphics[width=1in,height=1.25in,clip,keepaspectratio]{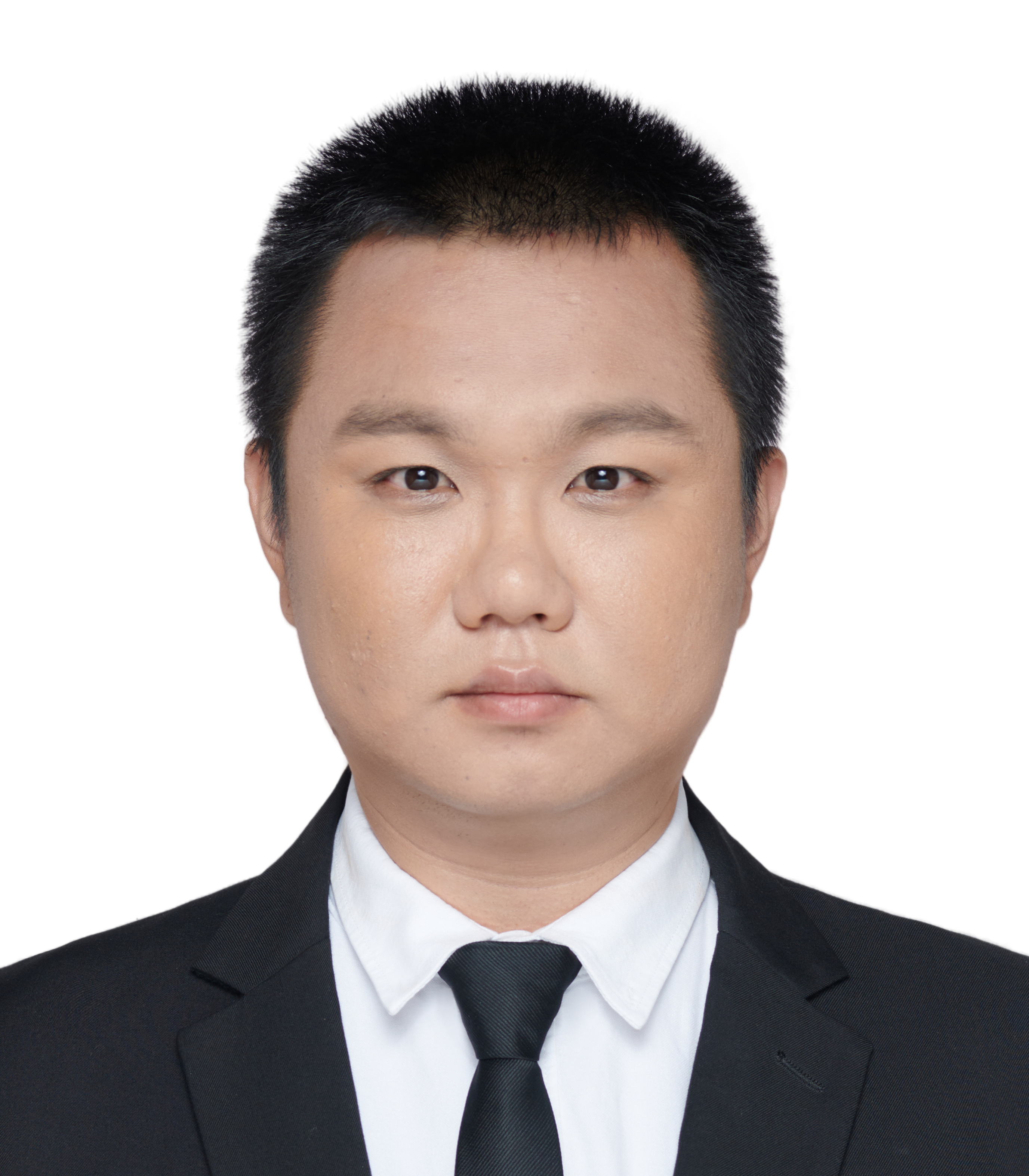}}]{Renqiang Luo}
  received the B.Sc. degree from University of Science and Technology of China, Hefei, China, in 2016, and the M.Sc. degree from University of South Australia, Adelaide, Australia, in 2019. He received a Ph.D. degree in the School of Software, Dalian University of Technology, Dalian, China, in 2024. Dr. Renqiang Luo is currently an Assistant Professor in the Jilin University, Changchun, China. His research interests include graph learning, algorithmic fairness, and trustworthy AI.
\end{IEEEbiography}

\vspace{-12pt}
\begin{IEEEbiography}[{\includegraphics[width=1in,height=1.25in,clip,keepaspectratio]{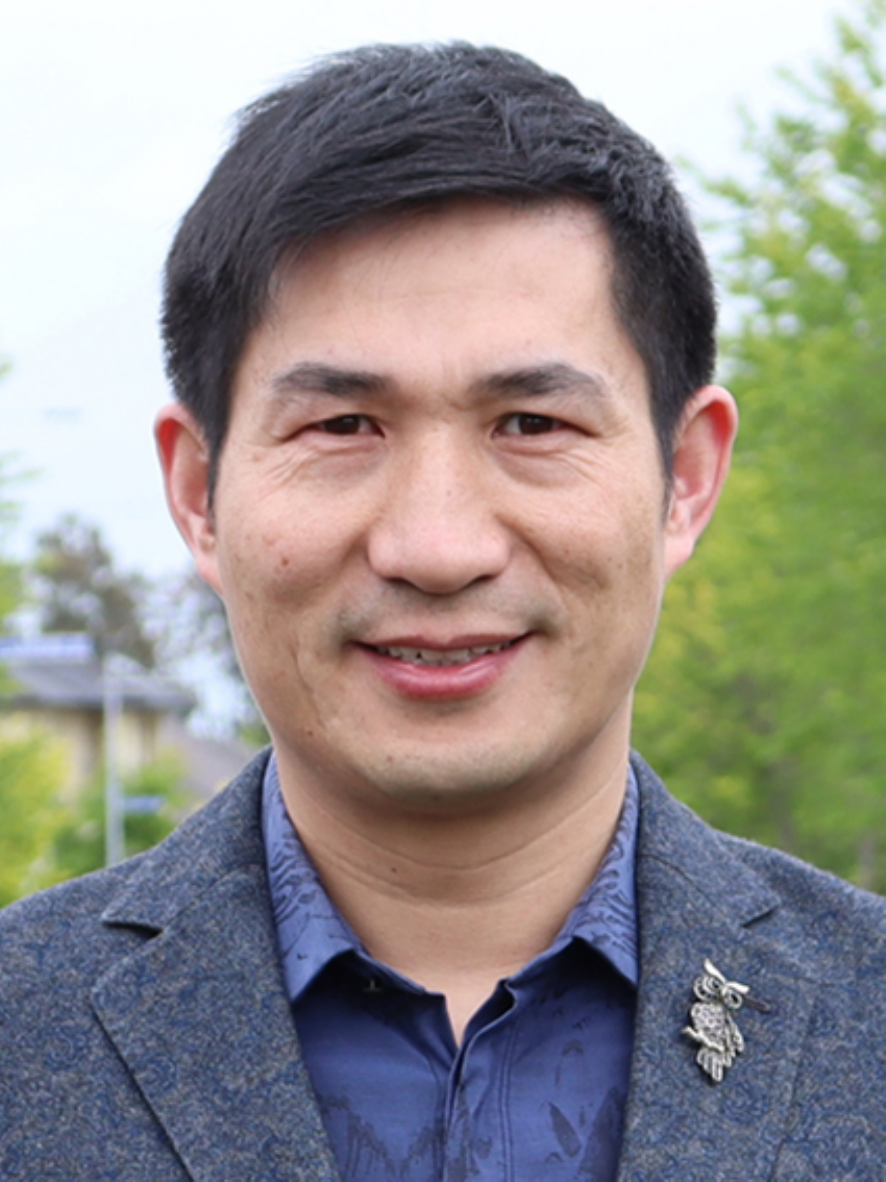}}]{Feng Xia}
(Fellow, IEEE) received the BSc and PhD degrees from Zhejiang University, Hangzhou, China. He is a Professor in School of Computing Technologies, RMIT University, Australia. Recognized as a Clarivate Highly Cited Researcher and a ScholarGPS Highly Ranked Scholar, Dr. Xia has published over 400 scientific papers. His work is featured in top-tier journals and conferences. Dr. Xia has extensive editorial and organizational experience, having served as an Associate or Guest Editor for over 20 journals and in various Chair roles for more than 30 conferences. His contributions and leadership have been recognized by prestigious awards. He has delivered numerous keynote speeches and invited talks at international venues worldwide. He is the Chair of IEEE Task Force on Learning for Graphs. His research interests include artificial intelligence, graph learning, brain, robotics, and cyber-physical systems. He is a Fellow of the IEEE.
\end{IEEEbiography}

\end{document}